%% file: bare_jrnl_new_sample4.tex
\documentclass[lettersize,journal]{IEEEtran}
\usepackage{amsmath,amsfonts}
\usepackage{algorithmic}
\usepackage{algorithm}
\usepackage{array}
\usepackage[caption=false,font=normalsize,labelfont=sf,textfont=sf]{subfig}
\usepackage{textcomp}
\usepackage{stfloats}
\usepackage{url}
\usepackage{verbatim}
\usepackage{graphicx}
\usepackage{cite}
\usepackage{enumitem}
\usepackage{hyperref}
\usepackage{booktabs}%
\usepackage{float}
\usepackage{makecell}  
\usepackage{pifont}    
\usepackage{multirow}%
\usepackage{xcolor}%
\usepackage{caption}

\newcommand{\etal}[0]{\textit{et al.}}
\newcommand{\ourmodel}[0]{FaNO}
\newcommand{\ournetwork}[0]{FaNO}
\newcommand{\ourframework}[0]{UGF}

\begin{document}

\title{Factorized Neural Operators Decompose Dynamic and Persistent Responses}

\author{Hao Tang, Yuechen Duan, Jiongyu Zhu, Zimeng Feng, Hao Li, Chao Li%
\thanks{
Hao Tang is with the University of Dundee, U.K.
Yuechen Duan, Jiongyu Zhu, Zimeng Feng and Hao Li are with Fudan University, China.
Chao Li is with the University of Dundee, U.K., and also with the University of Cambridge, U.K.
Hao Tang and Yuechen Duan contributed equally.

Corresponding author: Chao Li (e-mail: cl647@cam.ac.uk).
}
}

\markboth{Journal of \LaTeX\ Class Files,~Vol.~14, No.~8, August~2021}%
{Shell \MakeLowercase{\textit{et al.}}: A Sample Article Using IEEEtran.cls for IEEE Journals}


\maketitle

\begin{abstract}
Physical systems often exhibit heterogeneous mechanisms, where rapidly evolving dynamics coexist with persistent structures.
Capturing such multiscale physical behavior remains challenging for existing neural operators, which typically rely on single dominant inductive bias and therefore couple distinct physical responses into a shared representation.
We introduce the Unified Green's Function Framework across domains and propose the Factorized Neural Operators (FaNO), which decompose spectral representations into equivariant-inspired dynamic responses and invariant-inspired persistent responses, leading to better interpretability and generalization.
Mechanistically, we show that the two operator branches spontaneously specialize into distinct physical roles that remain consistent across scales and domains: the equivariant-inspired branch captures rapidly varying transient dynamics, whereas the invariant-inspired branch extracts coherent persistent structures.
This factorized mechanism of FaNO consistently improves prediction accuracy, parameter efficiency and cross-scale generalization across physical systems and domains.
In particular, it maintains consistent predictions under long-horizon autoregressive rollout, cross-resolution extrapolation and physical-regime shifts. 
These findings suggest that scalable physical modeling may benefit from moving beyond single-inductive-bias formulations toward factorized operator representations that better reflect the heterogeneous organization of physical systems, accelerating the reliable deployment of machine learning for scientific computing and discovery.
\end{abstract}


\section{Introduction}
\IEEEPARstart{A}{ccurately} modeling complex physical systems is a central objective of scientific computing. Systems such as geophysical flows, weather and climate dynamics, engineering turbulence, wave propagation and biological surfaces involve interactions across multiple spatial and temporal scales~\cite{bonev2023spherical,liu2024evaluation,hu2025spherical,tang2026generalized,zhao2021spherical}. A recurring challenge is that they rarely exhibit a single homogeneous response. Rapidly evolving dynamics often coexist with persistent structures induced by geometry, boundary conditions, and heterogeneous media~\cite{weinan2007heterogeneous,peng2021multiscale,lucarini2024detecting}.
Such heterogeneous responses often exhibit distinct sensitivity to discretization, temporal evolution, and domain geometry, making it difficult for computational models to reliably capture both transformation-sensitive dynamics and transformation-stable structure, while remaining robust across resolution, physical regimes, and computational domains~\cite{gao2025discretization,li2020neural,kovachki2023neural}.

Neural operators offer a principled framework by learning mappings between function spaces and approximating solution operators of partial differential equations (PDEs)~\cite{kovachki2023neural}. Unlike conventional neural networks trained on fixed discretizations~\cite{lecun1998gradient,ronneberger2015u,hornik1989multilayer,he2016deep}, they can support discretization-agnostic inference and rapid surrogate simulation~\cite{lutjens2022multiscale}. Spectral neural operators based on Fourier, spherical and manifold harmonics are particularly attractive because they encode global interactions efficiently and exploit geometry-native bases~\cite{li2020fourier,bonev2023spherical,tang2026generalized,giambagli2021machine,chen2024learning}. Yet, most existing spectral operators impose a single dominant inductive bias on the full response~\cite{han2022geometrically,finzi2021residual,worrall2019deep,cohen2016group}. 
This is limiting for physical systems that couple heterogeneous components: the same mechanism must account simultaneously for rapidly varying dynamics and for persistent, system-dependent structures that are not well explained by equivariance alone~\cite{finzi2021residual,zheng2024relaxing,wang2022approximately,wad2022equivariance}. 
As a result, such response coupling can reduce interpretability, amplify autoregressive errors and weaken cross-resolution generalization, especially in physical systems whose mechanisms evolve over widely separated spatial and temporal scales~\cite{weinan2007heterogeneous}.

This challenge raises a key question in neural operator design: should physical operators be learned as a single response, or can they be factorized into mechanism-specific components? 
Classical Green’s-function formulations provide a natural perspective. A Green’s function serves as an integral response kernel to represent the solution of a physical system, whose structure reflects how information propagates under geometry and boundary constraints~\cite{von1984computational,pan2019green,einstein1934method,tang2026generalized}. Transformation-dependent kernels naturally lead to dynamic responses, whereas absolute position-dependent kernels can represent persistent structures that remain stable under transformations. 
This view motivates assigning different components of a learned operator to different physical roles. However, classical Green’s functions are typically fixed and derived for specified operators, domains and boundary conditions, which limits their direct use in data-driven settings where physical responses arise from multiple interacting mechanisms. 
We therefore reinterpret Green’s-function not as an exact analytic kernel to be recovered, but as a design principle for geometry-native operator systems. This reformulation allows distinct components to encode different response mechanisms while enhancing the spectral efficiency and discretization robustness of neural operators.

Here we introduce a unified Green’s-function framework for neural operator design and propose Factorized Neural Operators (\ourmodel), a spectral architecture that decomposes physical responses into dynamic and persistent components. 
The dynamic branch models transformation-sensitive responses through an equivariant spectral operator. The persistent branch models stable, system-dependent responses through an invariant spectral operator driven by geometry-aware integral information. By separating symmetry interactions from channel mixing, FaNO preserves the efficiency and grid-invariance of spectral neural operators while enabling explicit branch specialization across Euclidean grids, spherical domains and geometric manifolds.

We evaluate FaNO across spherical, Euclidean and unstructured geometric domains spanning geophysical forecasting, canonical PDE benchmarks and real-world surface learning. Across these settings, response factorization improves predictive accuracy, parameter efficiency, cross-resolution generalization and long-horizon rollout stability. The two branches also exhibit consistent functional specialization: the dynamic component captures rapidly varying, sample-dependent structures, whereas the persistent component encodes stable, system-dependent spatial patterns. These findings support factorized operator representations as a general computational principle for modeling heterogeneous physical systems, moving neural operator design beyond single-branch dynamic learning toward mechanism-specific response modeling.

Our contributions are summarized as follows:
\begin{enumerate}[leftmargin=*]

\item We present a designable Green's-function framework for neural operator learning that unifies Euclidean, spherical and manifold domains under an operator formulation.

\item We propose Factorized Neural Operators (\ourmodel), which explicitly decompose dynamic and persistent responses through factorized operator representations, moving beyond the single-inductive-bias assumption while preserving spectral efficiency and discretization invariance.

\item We demonstrate that response factorization of \ourmodel~naturally induces mechanism specialization, where the dynamic branch captures transient behaviors and the persistent branch models coherent structures, leading to improved interpretability, long-horizon stability and cross-scale generalization in geophysics and fluid dynamics.

\item We conduct extensive evaluations across physical systems, geometries and scales, showing that \ourmodel~consistently improves performance, robustness and parameter efficiency.

\end{enumerate}

This paper builds on the previous conference version, which focused on spherical Green's-function parameterization for spherical neural operators~\cite{tang2026generalized}. In contrast, the present work introduces a unified factorized operator framework across Euclidean, spherical and manifold domains, together with explicit response decomposition, comprehensive mechanism analysis and significantly expanded empirical validation across diverse physical systems and domains.

\section{Related Work}
\label{sec:related_work}

\subsection{Neural operator learning} 
Neural operators enable efficient modeling for partial differential equations and scientific simulations~\cite{kovachki2023neural}.
Branch--trunk neural operators approximate nonlinear operators through separable function representations, especially deep operator network (DeepONet)~\cite{lu2021learning} and its variants~\cite{xiao2025quantum,he2024geom,he2023novel,goswami2023physics,zhu2023fourier}. 
Graph-based neural operators learn parametric kernels on irregular discretizations through graph message passing~\cite{li2020neural,li2020multipole,li2025harnessing,li2023geometry}.
Attention-based neural operators replace kernel integral operators with attention mechanisms to improve flexibility on irregular observations and arbitrary query locations~\cite{cao2021choose,hao2023gnot,wu2024transolver,nguyen2023climax,pathak2022fourcastnet}.
In contrast, spectral-based neural operators enable more efficient global modeling while preserving the intrinsic geometry of the underlying computational domain. Especially, Fourier Neural Operator (FNO) and its variants achieve spectral-domain parameterization of integral operators on structured grid~\cite{li2020fourier,tran2021factorized,duruisseaux2025fourier,guibas2021adaptive,liu2023domain,cao2024laplace}. 
Spherical harmonic-based approaches generalize Fourier-based operator learning to spherical fields and achieved high performance on geophysical simulations and weather prediction~\cite{bonev2023spherical,hu2025spherical,mahesh2024huge1,mahesh2024huge2,tang2026generalized}.
Recently, spectral operator learning has also been extended to irregular geometries. Neural Operators on Riemannian Manifolds (NORM)~\cite{chen2024learning,tang2025geometric} generalize operator learning to manifold domains through intrinsic eigenbases.

\subsection{Inductive bias and heterogeneous modeling}

Inductive bias plays a central role in modern scientific machine learning by incorporating known physical structures into learning algorithms. Among various inductive biases, equivariance has become one of the most successful principles for geometric representation learning. Group-equivariant convolutional networks~\cite{cohen2016group,worrall2017harmonic,cohen2018spherical,weiler2019general,cohen2019gauge,finzi2020generalizing} have demonstrated that embedding transformation symmetries into network architectures can substantially improve efficiency, robustness, and generalization. Similar ideas have also been adopted in neural operator learning, where Fourier-based operators naturally induce geometric equivariance through the corresponding spectral representations~\cite{li2020fourier,bonev2023spherical}.
Recognizing that strict equivariance may not always be sufficient for complex physical systems, recent studies have explored relaxed equivariant convolution for more expressive modeling through residual pathways, positional encoding, approximate equivariance, learnable symmetry breaking~\cite{finzi2021residual,wang2022approximately,zheng2024relaxing,wu2025relaxed,wu2025r2det,li2020fourier,bonev2023spherical,tang2026generalized}.
However, these methods remain mechanism-agnostic: heterogeneous physical responses, such as rapidly evolving dynamics, coherent large-scale and geometry-aware structures, are modeled within the latent representation rather than being explicitly associated with distinct physical mechanisms~\cite{weinan2007heterogeneous,peng2021multiscale,lucarini2024detecting,wang2022approximately,zheng2024relaxing}. Overall, explicitly representing such heterogeneous physical mechanisms within a unified operator framework remains challenging (Fig.~\ref{fig:model}A \& B), which is crucial for the interpretability, robustness, and generalization of operator learning.

\input{methods}

\input{results}

\section{Conclusion}
\label{Conclusion}

This work addresses a challenging question in operator learning: should heterogeneous physical systems be represented through a single shared operator response, or through multiple components associated with distinct physical mechanisms? Across a broad range of physical systems and computational domains, our results consistently support the latter view. 
By separating dynamic and persistent responses, \ourmodel\ provides a more suitable inductive bias for modeling physical systems whose behaviors emerge from the interaction of transient dynamics and long-lived structures.

The proposed framework is motivated by a unified Green's-function perspective, where operator responses are not required to share the same transformation behavior. This leads naturally to a factorized representation consisting of equivariant-inspired dynamic and invariant-inspired persistent components in \ourmodel. A central observation throughout our experiments is that this decomposition is not merely architectural: the two branches spontaneously specialize into distinct physical roles without explicit supervision. The dynamic branch captures rapidly evolving, state-dependent responses, whereas the persistent branch extracts coherent structures associated with geometry, boundary conditions, and large-scale physical organization. Together, these findings suggest that heterogeneous physical systems may themselves be organized through multiple response mechanisms, which can be reflected directly in operator design.

This factorized mechanism helps explain the consistent improvements observed across scales, domains and physical regimes. FaNO remains robust under long-horizon rollout, cross-resolution extrapolation and distribution shifts, while generalizing across geophysical systems, fluid dynamics, time-independent physical modeling and geometric learning. Importantly, these improvements are achieved through controlled architectural modifications rather than increased model capacity, indicating that the benefits arise from the factorized operator principle itself.
More broadly, our findings suggest a limitation of equivariance-only formulations for physical operator learning. Equivariance remains a powerful inductive bias for modeling transformation-consistent dynamics, but many physical systems also contain persistent structures that are not fully characterized by equivariant interactions alone. The results therefore motivate a broader perspective in which different physical responses may require different transformation behaviors within a unified operator framework.

Overall, this work suggests that scalable physical modeling may benefit from moving beyond single-inductive-bias operator formulations toward factorized representations that better reflect the heterogeneous organization of physical systems. By connecting physical insight, operator theory and neural operator design, factorized neural operators provide a step toward more interpretable, robust and generalizable foundations for scientific computing and physical discovery.

\textit{Limitations and Future directions:}
The current formulation represents one possible realization of the proposed principle. In addition, the present dynamic–persistent factorization primarily targets one form of physical heterogeneity. Other settings, including coupled multiphysics interactions and more complex physical constraints, may require alternative Green’s-function constructions and richer response decompositions. Exploring these possibilities represents an important direction for future work in Physics-informed Machine Learning.

\section*{Acknowledgements}

Chao Li declares the support from Guarantors of Brain.

\section*{Competing interests}

The authors declare no competing interests.

\bibliographystyle{IEEEtran}
\bibliography{ref}

\newpage

\end{document}

%% file: methods.tex
\section{Methodology}\label{sec11}

\begin{figure*}[t]
    \centering
    \includegraphics[width=0.95\linewidth]{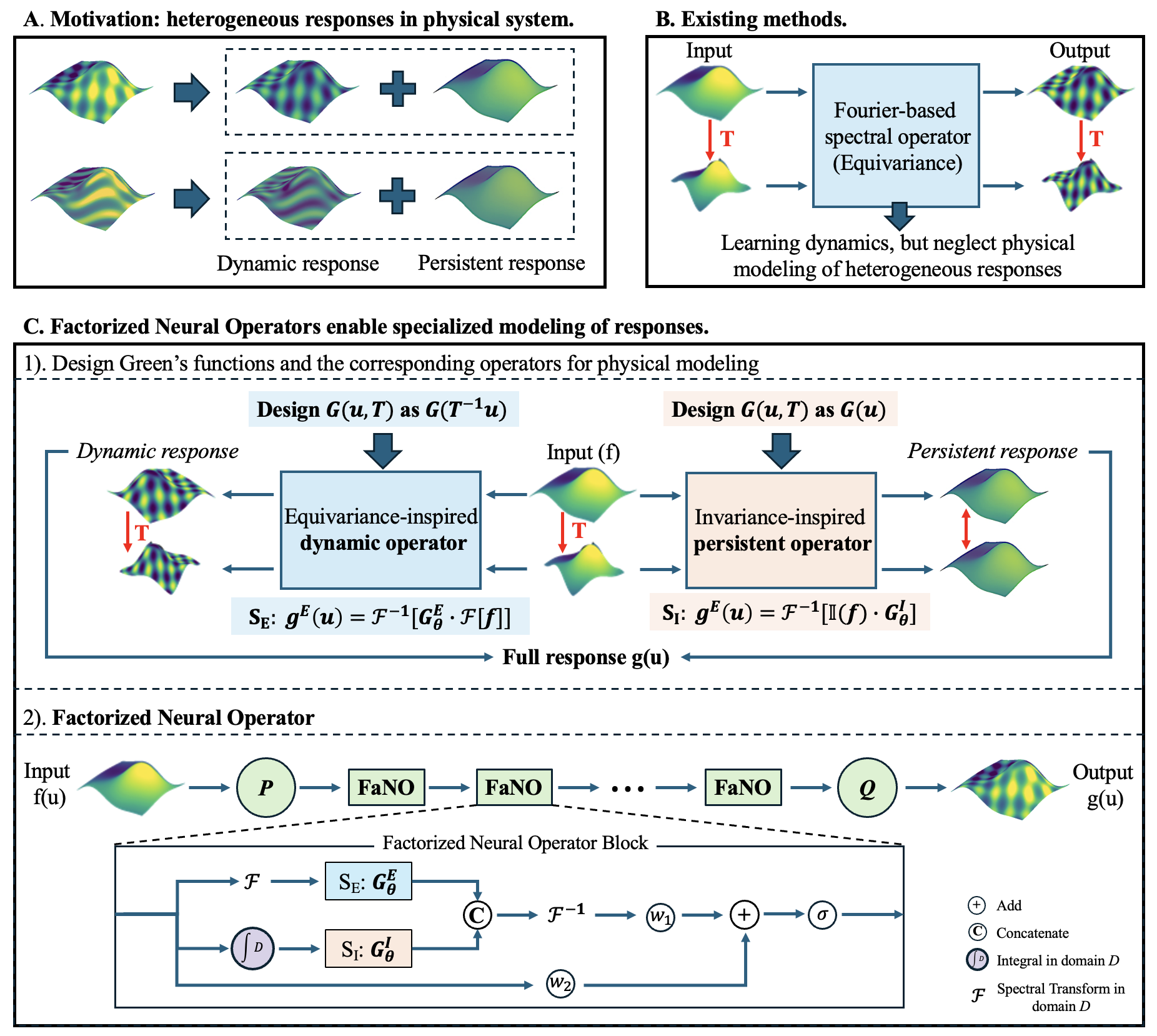}
    \caption{The motivation and model architecture of \ourmodel. 
    Physical systems often exhibit heterogeneous responses (\textbf{A}) while existing spectral operators tend to couple these responses into the latent representation (\textbf{B}). FaNO addresses this by factorizing the response into dynamic and persistent operator components and then fusing them into the full prediction. (\textbf{C.1}), resulting in the Factorized Neural Operator model (\ourmodel) (\textbf{C.2}), where $\mathcal{F}$ and $\mathcal{F}^{-1}$ represent spectral transform and inverse transform in domain $D$; $w$ is the linear transform layer and $\sigma$ is the activation function. $\textit{P}$ and $\textit{Q}$ are the encoder and decoder for channel transformation.}
    \label{fig:model}
\end{figure*}

The Green's function method provides a classical framework for solving linear partial differential equations (PDEs), where the solution is represented as an integral operator whose kernel is the Green's function~\cite{li2020neural}. The underlying design of Green's function determines how physical responses propagate and therefore provides a principled mechanism for designing inductive biases in operator learning~\cite{tang2026generalized}. This motivates designing different Green's functions to simulate different response mechanisms beyond existing Fourier-based design (Fig.~\ref{fig:model}A \& B).

In this section, we first introduce a \textbf{Unified Green's Function framework} (UGF) that provides a unified formulation for designable Green's functions across Euclidean, spherical, and manifold domains (Sec.~\ref{subsec:UGF}). Based on UGF, we derive the proposed \textbf{dynamic} and \textbf{persistent} operators, corresponding to complementary physical response mechanisms (Fig.~\ref{fig:model}C in Sec.~\ref{subsec:2O2G}). Finally, the two operators are integrated through a lightweight factorized architecture to construct the proposed \textbf{Factorized Neural Operator (FaNO)} (Fig.~\ref{fig:model}D in Sec.~\ref{subsec:FaNO}).

\subsection{Unified Designable Green's Function Framework}
\label{subsec:UGF}

We first present a unified formulation that is independent of specific geometric structure such as group symmetries.
Let \(D\) denote a general domain, such as the Euclidean space \(\mathbb{R}^{d}\), the sphere \(S^{2}\), or a more general manifold. Let \(\mathcal{O}\) be a linear differential operator defined on \(D\), and consider the PDE:
\begin{equation}
\mathcal{O}[g(u)] = f(u), \qquad u \in D,
\label{eq:pde_general}
\end{equation}
where \(f(u)\) is the input function and \(g(u)\) is the target solution.
The Green's function is defined as a two-point kernel \(G(u,v)\), satisfying
\begin{equation}
\mathcal{O}_{u} [G(u,v)] = \delta(u,v), \qquad u,v \in D,
\label{eq:green_general}
\end{equation}
where \(\delta(u,v)\) denotes the Dirac delta distribution on \(D\), and \(\mathcal{O}_{u}\) indicates that the operator acts on the variable \(u\).
Under this definition, the solution to Equation~\eqref{eq:pde_general} can be written in the unified integral form
\begin{equation}
g(u) = \int_{D} G(u,v)\, f(v)\, dV(v),
\label{eq:solution_general}
\end{equation}
where $dV(v)$ denotes the appropriate measure on \(D\).
To verify Equation~\eqref{eq:solution_general}, we apply the operator \(\mathcal{O}_{u}\):
\begin{align}
\mathcal{O}_{u} [g(u)]
&= \mathcal{O}_{u} \left[ \int_{D} G(u,v)\, f(v)\, dV(v) \right] \nonumber \\
&= \int_{D} \mathcal{O}_{u} [G(u,v)]\, f(v)\, dV(v) \nonumber \\
&= \int_{D} \delta(u,v)\, f(v)\, dV(v) \nonumber \\
&= f(u),
\label{eq:DGF_proof}
\end{align}
which verifies that \(g(u)\) is the solution of Equation~\eqref{eq:pde_general}.

\paragraph{Unified designable perspective}
The formulation in Equation~\eqref{eq:solution_general} provides a unified operator view of Green's functions that applies to different geometric domains beyond our prior spherical setting~\cite{tang2026generalized}. Based on this perspective, we propose the Unified Designable Green's Function Framework (\ourframework), which models general mappings \(f(u) \rightarrow g(u)\) by designing specific Green's functions $G(u,v)$ to simulate the physical systems and simplify Equation~\eqref{eq:solution_general} and derive the corresponding operator solution $g(u)$. 

\paragraph{\ourframework~in structured domains}
In structured domains, e.g., Euclidean space $\mathbb{R}^{d}$ and sphere $S^{2}$, additional geometric structure allows further simplification of the above formulation. Specifically, when the domain \(D\) admits a transitive group action \(\mathcal{G}\), any point \(v \in D\) can be represented as \(v = T u_0\), where \(u_0\) is a fixed reference point and \(T \in \mathcal{G}\). In this case, the Green's function can be equivalently expressed as \(G(u,T)\), and Equation~\eqref{eq:solution_general} becomes:
\begin{equation}
g(u) = \int_{\mathcal{G}} G(u,T)\, f(Tu_{0})\, d\mu(T),
\label{eq:solution_group_origin}
\end{equation}
where \(d\mu(T)\) denotes the invariant measure on \(\mathcal{G}\).
This formulation recovers several important structured cases. In \(D=\mathbb{R}^{d}\), \(\mathcal{G}\) corresponds to the translation group. On the sphere \(D=S^{2}\), \(\mathcal{G}=SO(3)\) corresponds to rotations. More generally, \(\mathcal{G}\) may encode combinations of rotations and translations depending on the geometry of \(D\).

The specific process of \ourframework~in structured domains: (1) Design system-dependent $G(u, T)$, which constrains the system to be solved $\mathcal{O}$ by $\mathcal{O}_{u} [G(u,T)] = \delta(u,Tu_0) = \delta(T^{-1}u)$; $\rightarrow$ (2) Derive the system-constrained operator solution $g(u)$. 
Therefore, by designing specific Green's functions in the form \(G(u,T)\), the resulting operator can explicitly model how signals evolve under group transformations, which is important for physical systems, where underlying dynamics often exhibit structured variations governed by symmetries.

\subsection{Operators derived from Green's Functions}
\label{subsec:2O2G}
Under \ourframework, we design Equivariant Green's Function $G(T^{-1}u)$, 
Invariant Green's Function $G(u)$ and derive the corresponding \textbf{dynamic, persistent} operator solutions in structured domains with explicit symmetry, and then rigorously generalize the solution to unstructured geometric domain without explicit group structure.

\subsubsection{Dynamic Operators from $G(T^{-1}u)$}
\label{EO_G1}
Inspired by existing transformation-equivariant methods~\cite{bonev2023spherical,li2020fourier}, we design \( G(u,T) \) as \( G^E(T^{-1}u) \) based on strictly $\mathcal{G}$-symmetry to simulate rapidly varying dynamical system. Under this formulation, the dynamic operator solution \( g(u) \) is given by:
\begin{align}
\label{eq:sfno}
g(u) &= \int_{\mathcal{G}} G^E(T^{-1}u) f(Tu_0) \, d\mu(T).
\end{align}

The specific formulations of $\mathcal{F}[g(u)]$ in different domains are as follows (Detailed derivation is in Supplementary Material):

On the Euclidean domain $D_{\mathbb{R}^d}$, we design $G(u, v)$ as $G(u-v)$ based on the translational equivariance, therefore $\mathcal{F}[g(u)]$ is derived in $D_{\mathbb{R}^d}$ as:
\begin{align}
\label{eq:GE_solution_euclidean}
\mathcal{F}[g(u)] 
&= \mathcal{F}_{\mathbb{R}^d} [G^E](k) \cdot \mathcal{F}_{\mathbb{R}^d} [f](k),
\end{align}
where $\mathcal{F}_{\mathbb{R}^d}$ is the regular Fourier transform.

On the spherical domain $D_s$, we design $G(u, v)$ as $G(R^{-1}u)$ based on the rotational equivariance~\cite{driscoll1994computing} under the rotational group ($R$ from $SO(3)$), therefore $\mathcal{F}[g(u)]$ is derived in $D_s$ as:
\begin{align}
\label{eq:GE_solution_sphere}
\mathcal{F}[g(u)] 
&= {2\pi \sqrt{\frac{4\pi}{2l+1}}}\cdot \mathcal{F}_s [G^E](l, 0) \cdot \mathcal{F}_s [f](l, m),
\end{align}
where $\mathcal{F}_s$ is the spherical harmonic transform with the integer degrees \( l \geq 0 \) and orders \( |m| \leq l \)~\cite{muller2006spherical}.

On the general manifold domain $D_M$ without an explicit group structure, we design $G(u, v)$ as $\sum_{j=0}^{\infty} \hat{G}(j)\, \phi_j(u)\phi_j(v)$ based on the spectral diagonalizability and applying the Laplace--Beltrami operator on the Riemannian manifold~\cite{tang2025geometric, sharp2022diffusionnet, bronstein2017geometric}, therefore Equation~\ref{eq:solution_general} is derived and therefore $\mathcal{F}[g(u)]$ is rigorously extended on $D_M$ as:
\begin{align}
\label{eq:GE_solution_manifold}
\mathcal{F}[g(u)]
&= \mathcal{F}_M[G](i) \cdot \mathcal{F}_M[f](i),
\end{align}
where $\mathcal{F}_M$ denotes the spectral transform defined by the eigenfunctions of the Laplace--Beltrami operator $\Delta_g$, which generalize the structured Fourier basis to curved spaces. The eigenfunctions
$\{\phi_i\}_{i=0}^{\infty}$ satisfy
$\Delta_g \phi_i(u) = \lambda_i \phi_i(u)$, and form a complete orthonormal basis of $L^2(M)$.

\textbf{Unified dynamic operators:}
Thus, the dynamic operator solution $g^E(u)$ across domains is reconstructed and unified by:
\begin{equation}
g^E(u) = \mathcal{F}^{-1}[G^E_{\theta} \cdot \mathcal{F}[f]],
\label{eq:dynamic_solution}
\end{equation}
where \(G^E_{\theta}\) denote the learnable spectral weights parameterized by the neural operator and $\mathcal{F}$ is the spectral transform.

\begin{figure*}[ht]
\centering
\includegraphics[width=0.99\linewidth]{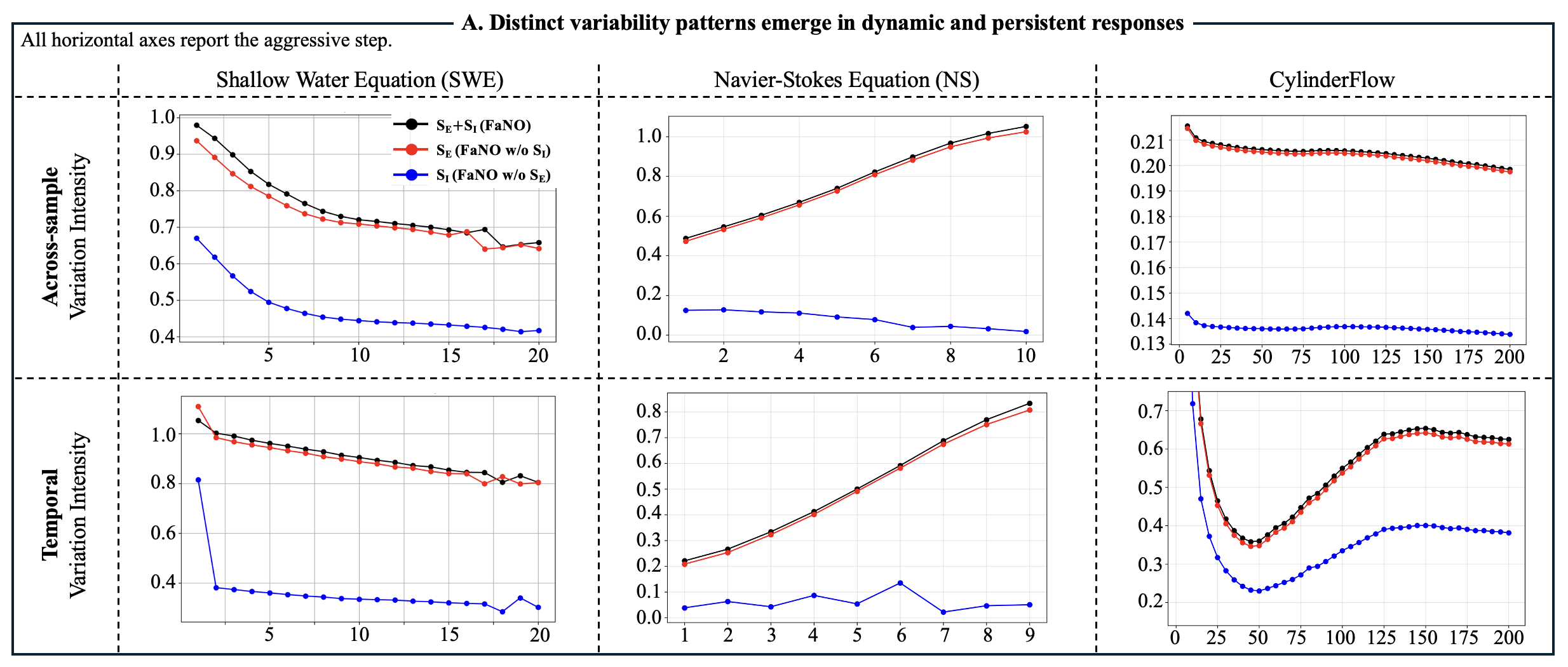}
\caption{Quantitative analysis of response heterogeneity in factorized operators across physical systems and domains: 
The response variation intensity across samples (top) and rollout horizons (bottom) for the full prediction ($\mathbf{S_E+S_I}$), the dynamic branch ($\mathbf{S_E}$), and the persistent branch ($\mathbf{S_I}$). 
Across domains, the dynamic branch maintains rapid evolution (high variation intensity), whereas the persistent branch maintains slower evolution (extremely low variation intensity) that remains stable across scales.}
\label{fig:quantitative_mechanism}
\end{figure*}

\subsubsection{Persistent Operators from $G(u)$}
\label{IO_G2}
While equivariant operators naturally capture dynamical responses that transform consistently with the input, many persistent structures remain largely unchanged under the same transformations.
Examples include large-scale circulation patterns in geophysical flows, background energy distributions in turbulence, and stable geometric structures on manifolds.
Modeling such persistent responses therefore requires operators that remain nearly invariant to these transformations.

To explicitly capture persistent responses that remain unchanged under some transformations, we design the Invariant Green's Function $G^I$ across domains:
\begin{align}
\label{G1u}
G(u,v) = G^I(u).
\end{align}
Under this formulation, the prediction target \( g(u) \) is given by:
\begin{align}
\label{eq:sfno}
g(u) &= \int_{\mathcal{G}} G^I(u) f(v) \, dV(v).
\end{align}

The specific formulations of $\mathcal{F}[g(u)]$ in different domains are as follows (Detailed derivation is in the Supplementary Material):

We derive the persistent operator solution in Euclidean domain under \ourframework~as:
\begin{align}
\label{eq:GI_solution_Euclidean}
\mathcal{F}_{\mathbb{R}^d}[g(u)] 
&= \int_{\mathbb{R}^d} \left( \int_{\mathbb{R}^d} G^I(u)\, f(v)\, dv \right) e^{-i k \cdot u}\, du \nonumber\\
&= C_f \cdot G_{\theta}^{I}(k),
\end{align}
where $C_f$ is the regular integral of the input function ($f(u)$). $G_{\theta}^{I}(k)$ is the learnable kernel.

We derive the persistent operator solution in spherical domain under \ourframework~as:
\begin{align}
\label{eq:GI_solution_sphere}
\mathcal{F}_s[g(u)] 
&= \int_{SO(3)} \left( \int_{S^2} G^I(u) f(Rn) \, dR \right) \overline{Y_l^m(u)} du \nonumber\\
&= C_f \cdot G_{\theta}^{I}(l, m),
\end{align}
where $C_f$ is the spherical integral of the input function ($f(u)$); $G_{\theta}^{I}(l, m)$ is the learnable and asymmetric kernel;
$Y_l^m(u)$ is the spherical harmonic function.

We further derive the persistent operator solution on Riemannian manifolds under \ourframework~as:
\begin{align}
\mathcal{F}_M[g(u)]
&= \int_M g(u)\, \phi_i(u)\, dV(u) \nonumber \\
&= \int_M \left( \int_M G^I(u)\, f(v)\, dV(v) \right)\phi_i(u)\, dV(u) \nonumber \\
&= C_f \cdot G_{\theta}^I(i).
\label{eq:GI_solution_manifold}
\end{align}
where $C_f$ is the geometry-aware integral of the input function ($f(u)$); $G_{\theta}^{I}(i)$ is the learnable kernel. \(dV(\cdot)\) is the Riemannian volume measure;

\textbf{Unified persistent operators:}
Thus, the persistent operator solution $g(u)$ across domains is reconstructed by:
\begin{equation}
g^I(u) = \mathcal{F}^{-1}[C_f \cdot G^I_{\theta}],
\label{eq:persistent_solution}
\end{equation}
where $C_f$ denotes the geometry-aware channel-wise spatial integral of the input function $f(u)$ and $\mathcal{F}^{-1}$ is the inverse spectral transform.
The invariant operators enforce that the model’s output or intermediate representations remain unchanged under a prescribed set of transformations, such as global rotations of the sphere.
This mechanism naturally captures persistent responses, which correspond to stable spatial structures whose overall statistics remain largely invariant under such transformations.
Rather than attempting to learn invariance from data, the invariant operators impose it by construction.
Together with equivariant operators that model transformation-consistent dynamical responses, the invariant operators provide a complementary mechanism for representing persistent structures in physical systems.
 
\subsection{Factorized Neural Operators and Implementation Detail}
\label{subsec:FaNO}

We decompose the symmetry-based operator solutions and channel interaction and fuse dynamic and persistent operators by concatenation (\text{C}) to explicitly integrate these heterogeneous responses in Equation~\ref{eq:dynamic_solution} \&~\ref{eq:persistent_solution}, resulting in the full Factorized Neural Operator (\ourmodel):
\begin{gather}
\label{eq:g(u)_summary_FaNO}
g(u) = \mathcal{F}^{-1} \left\{ \text{C}
[\mathcal{F}[f(u)] \cdot G^E_{\theta}, ~\mathbb{I}(f(u)) \cdot G^I_{\theta}]
\right\} \cdot w_1,
\end{gather}
where $w_1$ is a linear layer for channel interaction, $\mathbb{I}(f(u))$ is the geometry-aware integral of the input function $f(u)$ (Fig.~\ref{fig:model}C). Following the previous operator models~\cite{li2020fourier,bonev2023spherical,chen2024learning}, the linear layer $w_2$ is also added as the residual pathway in the full \ourmodel~block architecture shown in Fig.~\ref{fig:model}C 2). The full \ourmodel~network consists of multiple \ourmodel~blocks, encoder $P$ and decoder $Q$.

Equations.\ref{eq:g(u)_summary_FaNO} provides the full operator formulation of \ourmodel~across domains, following the corresponding intrinsic spectral transforms native to each geometry. 
Specifically, for Euclidean domains, the dynamic branch is implemented using truncated Fourier modes following FNO~\cite{li2020fourier}. The dynamic response is computed through spectral convolution in Fourier space, whereas the persistent branch applies the invariant operator using the global integral feature $\mathbb{I}_{\mathbb{R}^d}(f(u))$. The outputs of the two branches are concatenated and projected through a learnable linear layer. 
For spherical domains, the dynamic branch follows the spherical harmonic operator used in SFNO~\cite{bonev2023spherical}. Dynamic responses are represented through spherical harmonic coefficients and spectral filtering on the sphere. The persistent branch is constructed from the spherical integral feature $\mathbb{I}_{s}(f(u))$, which is multiplied by the learnable invariant kernel in spectral space before fusion with the dynamic response. 
For unstructured manifold domains, the dynamic branch is implemented using Laplace--Beltrami operator (LBO) following NORM~\cite{chen2024learning} while the persistent branch is generated from the global manifold integral feature $\mathbb{I}_{\mathbb{M}}(f(u))$. The two responses are fused in the spectral domain and mapped back to the spatial domain through the inverse LBO.

%% file: results.tex
\section{Experiments and Results}

\begin{table}[ht]
\centering
\caption{Matched single-branch spectral operators used for controlled architectural comparisons. For each benchmark, the listed operator serves as the corresponding single-branch counterpart of \ourmodel. All matched comparisons share identical backbone architectures, embedding dimensions, and training protocols, differing only in the operator block design.}
\label{tab:protocol}
\renewcommand{\arraystretch}{1.5}
\resizebox{\columnwidth}{!}{
\begin{tabular}{c!{\vrule}c!{\vrule}c!{\vrule}c}
\toprule
\textbf{Category} &
\textbf{Benchmark} &
\makecell{\textbf{Computational}\\\textbf{domain}} &
\makecell{\textbf{Single-branch}\\\textbf{counterpart}} \\
\midrule
\multirow{2}{*}{\makecell{Geophysical\\Modeling}}
& SWE & Sphere & SFNO~\cite{bonev2023spherical} \\
& WeatherBench & Sphere & SFNO~\cite{bonev2023spherical} \\
\midrule
\multirow{2}{*}{Fluid Dynamics}
& Navier--Stokes & Structured grid & FNO~\cite{li2020fourier} \\
& Cylinder flow & Geometric manifold & NORM~\cite{chen2024learning} \\
\midrule
\multirow{2}{*}{\makecell{Time-independent\\PDEs}}
& Darcy flow & Structured grid & FNO~\cite{li2020fourier} \\
& Helmholtz equation & Structured grid & FNO~\cite{li2020fourier} \\
\midrule
\multirow{3}{*}{\makecell{Geometric\\Learning}}
& Spherical MNIST & Sphere & SFNO~\cite{bonev2023spherical} \\
& RNA surface & Geometric manifold & NORM~\cite{chen2024learning} \\
& Human body & Geometric manifold & NORM~\cite{chen2024learning} \\
\bottomrule
\end{tabular}
}
\vspace{-0.3cm}
\end{table}

\begin{figure*}[hb]
\centering
\includegraphics[width=0.99\linewidth]{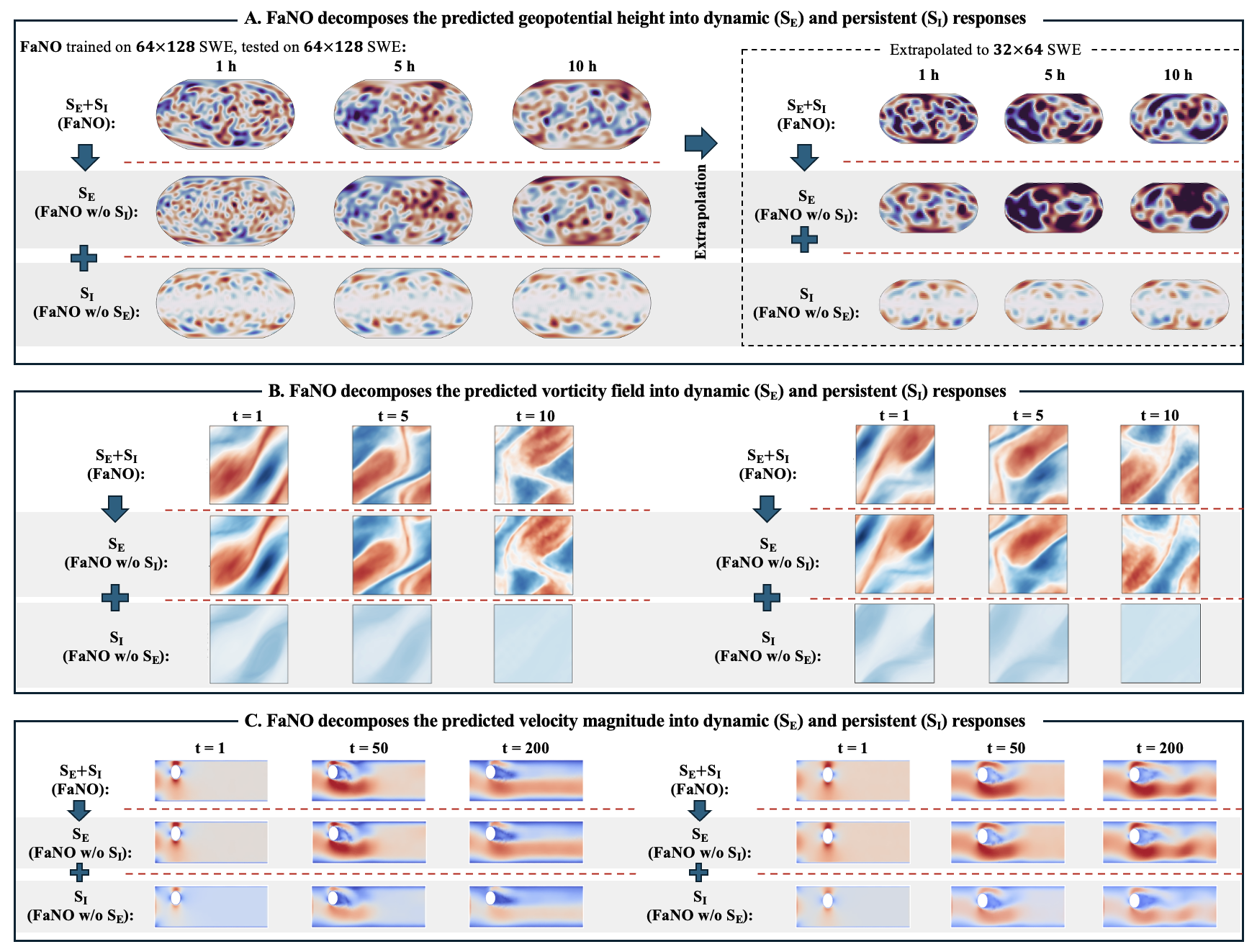}
\caption{Mechanism specialization of factorized operators across physical systems and domains:
Dynamic ($\mathbf{S_E}$) and persistent ($\mathbf{S_I}$) decomposition samples of:
\textbf{A}. geopotential height on spherical shallow water equations.
\textbf{B}. vorticity field in Navier--Stokes turbulence on Euclidean grids.
\textbf{C}. velocity field in cylinder wake prediction on unstructured meshes.
Across domains, the dynamic branch captures rapidly evolving transient responses, whereas the persistent branch extracts coherent large-scale structures that remain stable across scales.}
\label{fig:qualitative_mechanism}
\end{figure*}

\subsection{Experimental design overview}

Our experiments are designed to evaluate whether the response factorization provides a general computational principle for modeling heterogeneous physical systems. The experimental analysis proceeds from mechanism interpretation to progressively broader physical generalization. 

\textit{Experimental objectives:}
\textbf{First}, we investigate whether the proposed factorized operators spontaneously specialize into distinct physical roles across spherical, Euclidean and manifold domains, including geophysical flows, fluid turbulence and geometry-constrained dynamics. 
\textbf{Second}, we evaluate whether this mechanism improves multiscale geophysical modeling, and transfers to real-world Earth-system forecasting under changing spatial and time scales. 
\textbf{Third}, beyond geophysical systems, we further examine robustness under temporal, spatial and physical-regime shifts in canonical fluid dynamics, including cross-resolution extrapolation, long-horizon autoregressive rollout and viscosity-transfer evaluation. 
\textbf{Finally}, we evaluate whether the proposed factorization extends beyond fluid dynamics to time-independent physical systems and geometric learning tasks. 

\textit{Controlled Comparison and Ablation:} 
For direct architectural comparisons, \ourmodel~is evaluated against matched single-branch dynamic operator architectures, including SFNO~\cite{bonev2023spherical} on the sphere, FNO~\cite{li2020fourier} on the regular grid and NORM~\cite{chen2024learning} on the unstructured geometries under identical backbone architectures, optimization settings, differing only in the operator block design. 
These controlled comparisons therefore serve as direct architectural ablations for evaluating the role of response factorization, shown in the Table~\ref{tab:protocol}.
For broader benchmarking, we additionally compare with representative state-of-the-art baselines implemented under standard protocols. Implementation details are provided in the Supplementary Information.

Across all the spectral operator models, the total embedding dimension is kept unchanged.
In \ourmodel, the dynamic and persistent branches are assigned fixed channel ratios before concatenation, and the default setting uses an equal allocation of $0.5:0.5$ in this work.
Consequently, the concatenated representation has the same dimensionality as the existing single-branch operator, allowing \ourmodel\ to introduce only minimal computational overhead while significantly reducing the overall parameter count compared with existing single-branch spectral operators.
All experiments are implemented in PyTorch and conducted on NVIDIA V100 GPUs with 32GB memory.
More implementation details are provided in the Supplementary Information.

\textit{Data and Code Availability:} 
All datasets and benchmark implementations used in this work are publicly available from the official repositories:
WeatherBench~\cite{rasp2020weatherbench}
(\href{https://github.com/pangeo-data/WeatherBench}{GitHub}),
Shallow water equation~\cite{bonev2023spherical}
(\href{https://github.com/NVIDIA/torch-harmonics}{GitHub}),
Navier--Stokes and Darcy flow~\cite{li2020fourier}
(\href{https://github.com/li-Pingan/fourier-neural-operator}{GitHub}),
Cylinder Flow~\cite{pfaff2021meshgraphnets}
(\href{https://github.com/google-deepmind/deepmind-research/tree/master/meshgraphnets}{GitHub}),
Helmholtz~\cite{subramanian2023towards}
(\href{https://github.com/ShashankSubramanian/neuraloperators-TL-scaling}{GitHub}),
RNA and Human Mesh datasets~\cite{sharp2022diffusionnet}
(\href{https://github.com/nmwsharp/diffusion-net}{GitHub}),
and Spherical MNIST~\cite{cohen2018spherical}
(\href{https://github.com/jonkhler/s2cnn}{GitHub}).
Corresponding references and access information are also provided in the main text and Supplementary Information. The complete implementation of \ourmodel~is publicly available at: 
\url{https://github.com/haot2025/FaNO}.

\begin{figure*}[ht]
\centering
\includegraphics[width=1.0\linewidth]{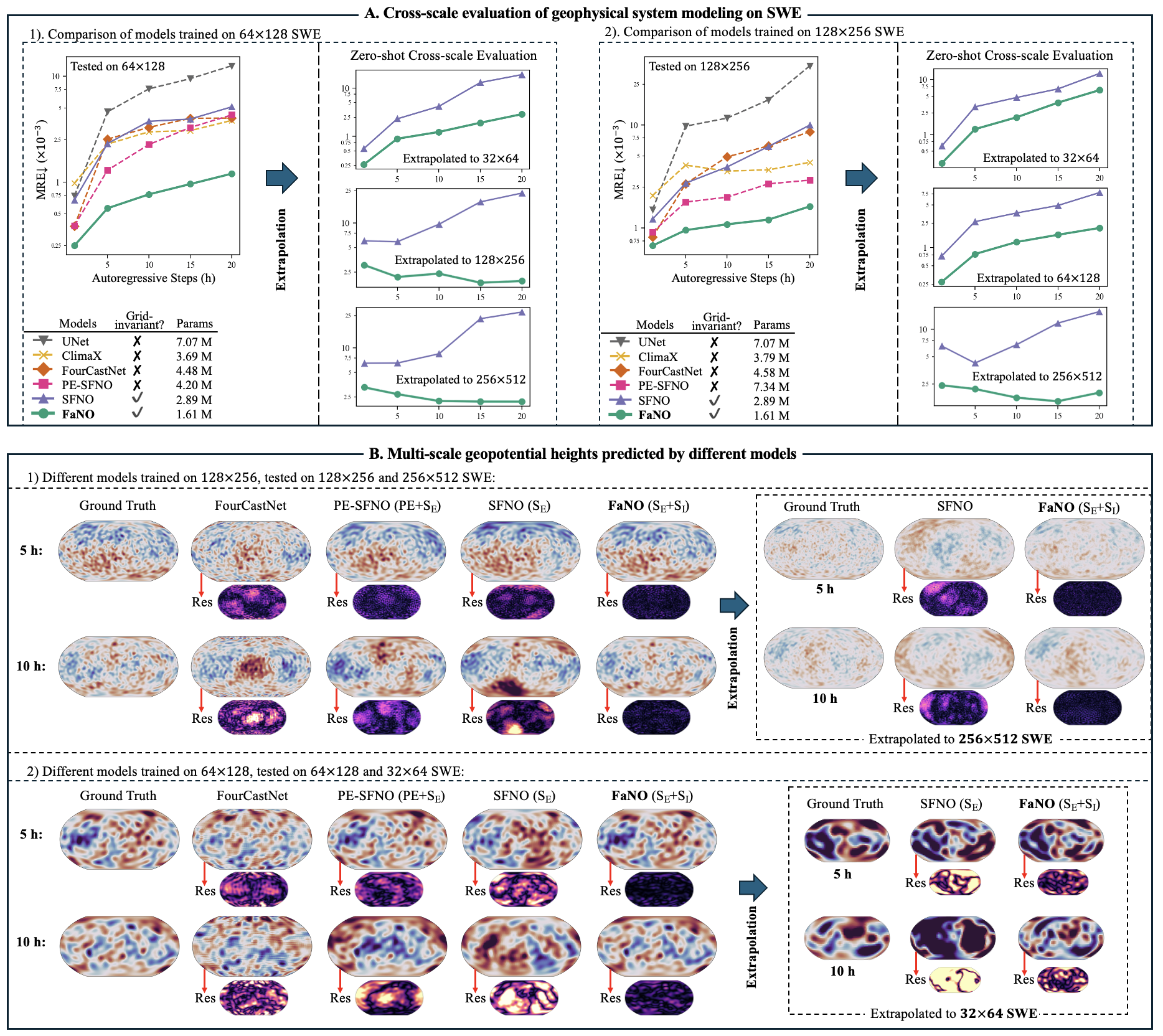}
\caption{Experimental results of multiscale geophysical system modeling on shallow water equations (SWE).
\textbf{A}. Mean relative error (MRE) comparison across temporal and spatial scales.
\textbf{B}. Geopotential-height prediction of different methods across temporal and spatial scales.}
\label{fig:swe}
\end{figure*}

\subsection{Factorized operator exhibits mechanism specialization across physical systems and domains}

We first investigate whether the two operator branches of \ourmodel~consistently organize into interpretable dynamic and persistent responses in representative physical systems.
The decomposition analysis is conducted in autoregressive forecasting settings across three physical systems and domains, including shallow water equations (SWE) on the sphere, Navier--Stokes turbulence (NS) on the regular grid, and geometry-constrained cylinder flow on the unstructured manifold (Details in the Supplementary Information). To isolate the physical role of each operator branch, we perform controlled branch interventions during inference. Specifically, the dynamic response ($\mathbf{S_E}$) is obtained by suppressing the persistent branch ($w/o~\mathbf{S_I}$), whereas the persistent response ($\mathbf{S_I}$) is obtained by suppressing the dynamic branch ($w/o~\mathbf{S_E}$). For autoregressive systems, the branch-suppressed predictions are computed along the same full-model rollout path, avoiding the confounding effect of accumulated rollout errors. This controlled decomposition allows us to directly examine the physical structures captured by each operator component.

To quantify the specialization behavior, Fig.~\ref{fig:quantitative_mechanism} reports the variability intensity across samples and rollout horizons (Details in Sec.~2B of Supplementary Information). A consistent pattern emerges across all systems: the variability of the dynamic branch closely matches that of the full prediction, indicating that most state-dependent changes are captured by $\mathbf{S_E}$. By contrast, the persistent branch exhibits substantially lower variability intensity across both samples and time, suggesting that it encodes structures that remain comparatively stable throughout system evolution.

Fig.~\ref{fig:qualitative_mechanism}A,B,C further present the decomposition behavior of \ourmodel\ across spherical, Euclidean, and manifold domains, including geopotential height on SWE, vorticity on NS, and velocity magnitude on geometry-constrained cylinder flow. Despite the substantial differences in geometry, discretization, and governing dynamics, a consistent specialization pattern emerges across all systems.

In all three systems, the dynamic branch $\mathbf{S_E}$ produces high-amplitude, spatially detailed responses that closely follow the full prediction $\mathbf{S_E}+\mathbf{S_I}$. 
These responses vary substantially across samples and rollout horizons, indicating that $\mathbf{S_E}$ is responsible for tracking transient, state-dependent and transformation-sensitive dynamics such as propagating waves in SWE, evolving vortical structures in NS, and time-dependent wake evolution behind the cylinder.

In contrast, the persistent branch $\mathbf{S_I}$ produces lower-amplitude but spatially coherent structures whose patterns vary only weakly across samples and time. 
In SWE, $\mathbf{S_I}$ concentrates on persistent polar and large-scale geopotential structures while remaining weak over the dynamically active mid-latitude regions. 
In NS, $\mathbf{S_I}$ highlights slowly varying coherent flow regions while suppressing rapidly changing vortical details. 
In cylinder flow, $\mathbf{S_I}$ focuses on the geometry-induced wake region downstream of the obstacle, especially the persistent flow band associated with the cylinder boundary.

These patterns suggest that $\mathbf{S_I}$ is not a second dynamic predictor competing with $\mathbf{S_E}$. Instead, it acts as a persistent structural anchor: a weakly varying component that preserves system-specific spatial organization induced by geometry, boundary conditions and large-scale physical constraints. The dynamic branch then models the dominant transient correction around this anchor. The full response combines these two roles, yielding predictions that retain both detailed dynamics and stable physical organization.
Importantly, the same specialization pattern appears in SWE, Navier--Stokes and cylinder flow despite their different domains, discretizations and governing physics. This cross-system consistency provides evidence that response factorization captures a general physical organization of operator learning.

\input{Tables/WB_5deg_ACC}

\begin{figure*}[hb]
\centering
\includegraphics[width=0.99\linewidth]{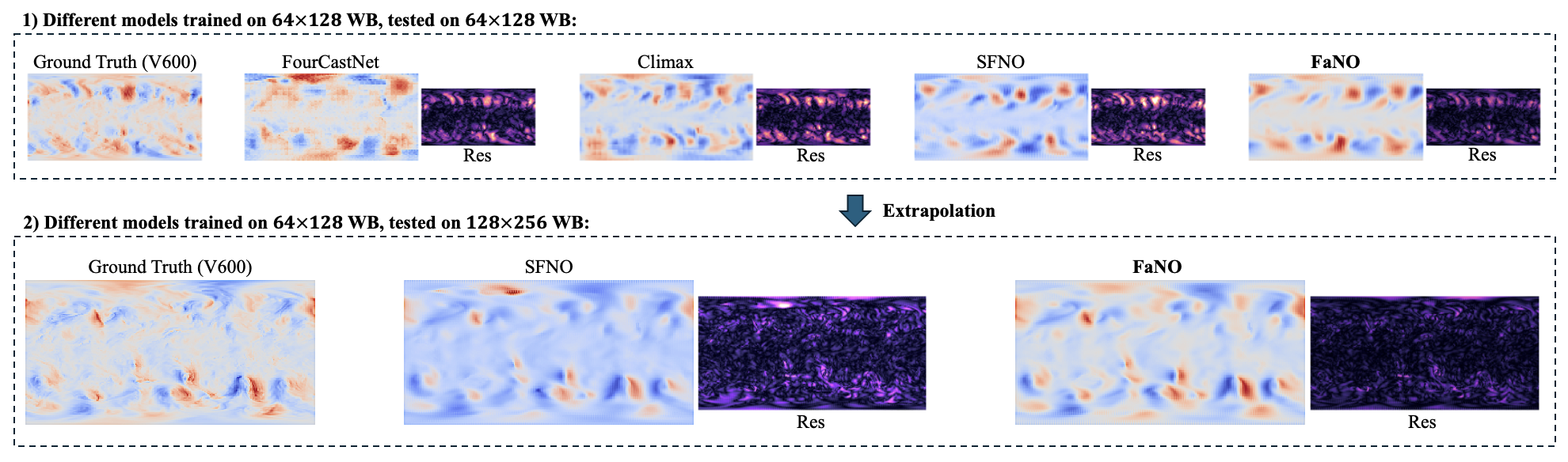}
\caption{Long-horizon prediction of wind velocity at 600 hPa pressure level (V600) at 120 h.}
\label{fig:weather}
\end{figure*}

\input{Tables/WB_2deg_ACC}

\subsection{Response factorization improves multiscale geophysical modeling}

We next evaluate whether the factorized mechanism of \ourmodel~observed in Fig.~\ref{fig:quantitative_mechanism} and~\ref{fig:qualitative_mechanism} translates into improved multiscale physical modeling under controlled geophysical dynamics. We consider the shallow water equations (SWE), the canonical model of large-scale geophysical fluid dynamics on the rotating sphere. Details are provided in the Supplementary Information.

Following~\cite{bonev2018discontinuous,bonev2023spherical}, we generate SWE trajectories using a classical spherical spectral solver~\cite{giraldo2001spectral}. Four datasets are constructed at spatial resolutions of $32\times64$, $64\times128$, $128\times256$, and $256\times512$, with a time step of $150$ s and three physical variables: geopotential height, vorticity, and divergence.
We compare \ourmodel\ with representative methods, including U-Net~\cite{ronneberger2015u}, ClimaX~\cite{nguyen2023climax}, FourCastNet~\cite{pathak2022fourcastnet}, SFNO~\cite{bonev2023spherical}, and SFNO with positional embedding (PE-SFNO)~\cite{bonev2023spherical}. All models are implemented under identical settings following prior spherical operator benchmarks~\cite{bonev2023spherical,tang2026generalized}: one-step training and autoregressive rollout. Additional implementation details are provided in the Supplementary Information.

Across all forecasting horizons and spatial resolutions shown in Fig.~\ref{fig:swe}A, \ourmodel\ consistently achieves the lowest prediction error while using only $1.61$M parameters, substantially fewer than competing methods. More importantly, the advantages become increasingly pronounced under cross-resolution evaluation, where models are directly transferred across unseen spatial discretizations.
\textbf{Under upward resolution extrapolation}, single-branch (equivariant-only) spectral baseline, SFNO exhibits severe degradation as rollout length increases, with prediction errors frequently approaching or exceeding $10^{-2}$ in MRE. By contrast, \ourmodel\ maintains stable prediction errors below $2.5\times10^{-3}$ across all forecasting horizons and resolution shifts, yielding more than a four-fold improvement over SFNO. 
\textbf{Under downward extrapolation}, SFNO rapidly accumulates autoregressive errors during long-term prediction and approaches $1\times10^{-2}$. In contrast, \ourmodel\ consistently maintains lower errors around $2.5\times10^{-3}$ or below and preserves more stable prediction quality under both spatial and temporal distribution shifts.

The qualitative comparisons in Fig.~\ref{fig:swe}B further support these observations. Existing methods accumulate visible phase and amplitude distortions during autoregressive rollout, especially under cross-resolution transfer. Error maps of SFNO become nearly saturated under downward extrapolation, indicating substantial loss of physical coherence. Under upward extrapolation, SFNO exhibits pronounced large-scale error concentrations and multiple localized failure regions. By contrast, \ourmodel\ preserves both large-scale circulation patterns and local dynamical structures with minimal degradation, maintaining coherent predictions even under unseen finer discretizations.

Taken together, these results suggest that response factorization ($\mathbf{S_E}+\mathbf{S_I}$) improves geophysical operator learning not only through lower prediction error, but also through substantially stronger cross-scale consistency under long-horizon rollout and changing discretizations.

\begin{figure}[h]
\centering
\includegraphics[width=0.88\linewidth]{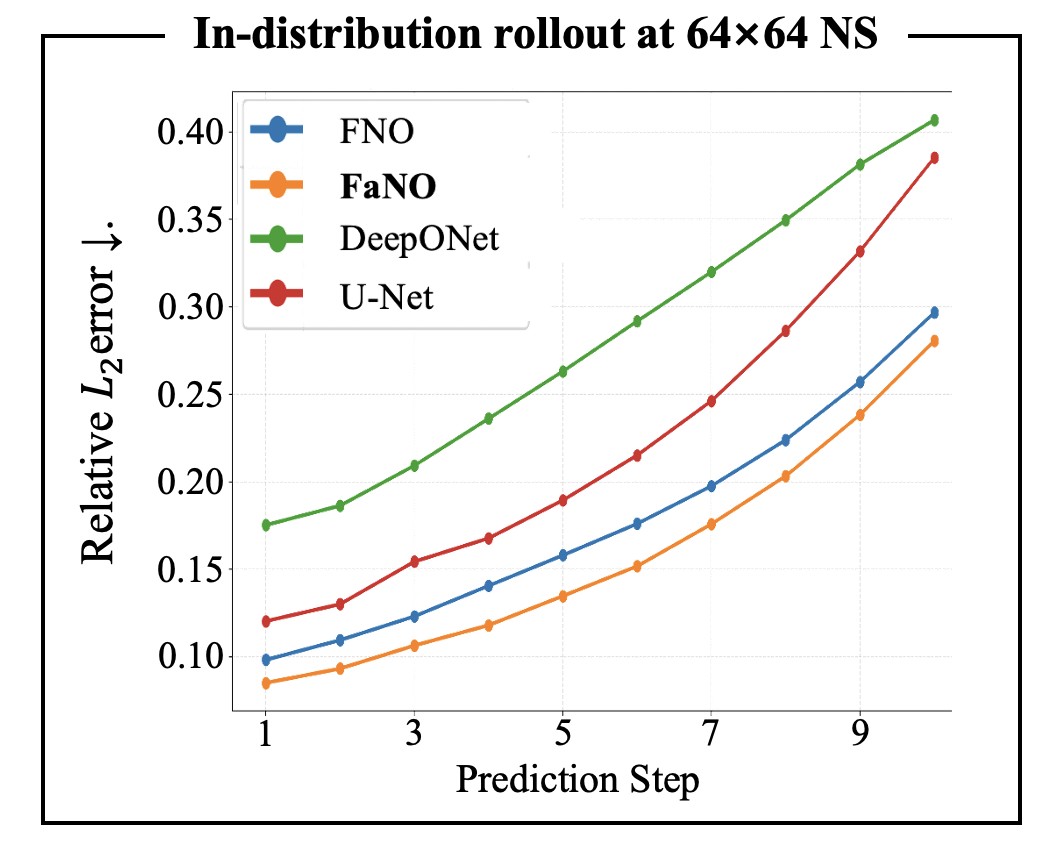}
\caption{Rollout evaluation of different methods on $64 \times 64$ Navier--Stokes turbulence (NS).}
\label{fig:NS_64}
\end{figure}

\subsection{Response factorization transfers to robust Earth-system forecasting}

\begin{figure*}[hb]
\centering
\includegraphics[width=0.90\linewidth]{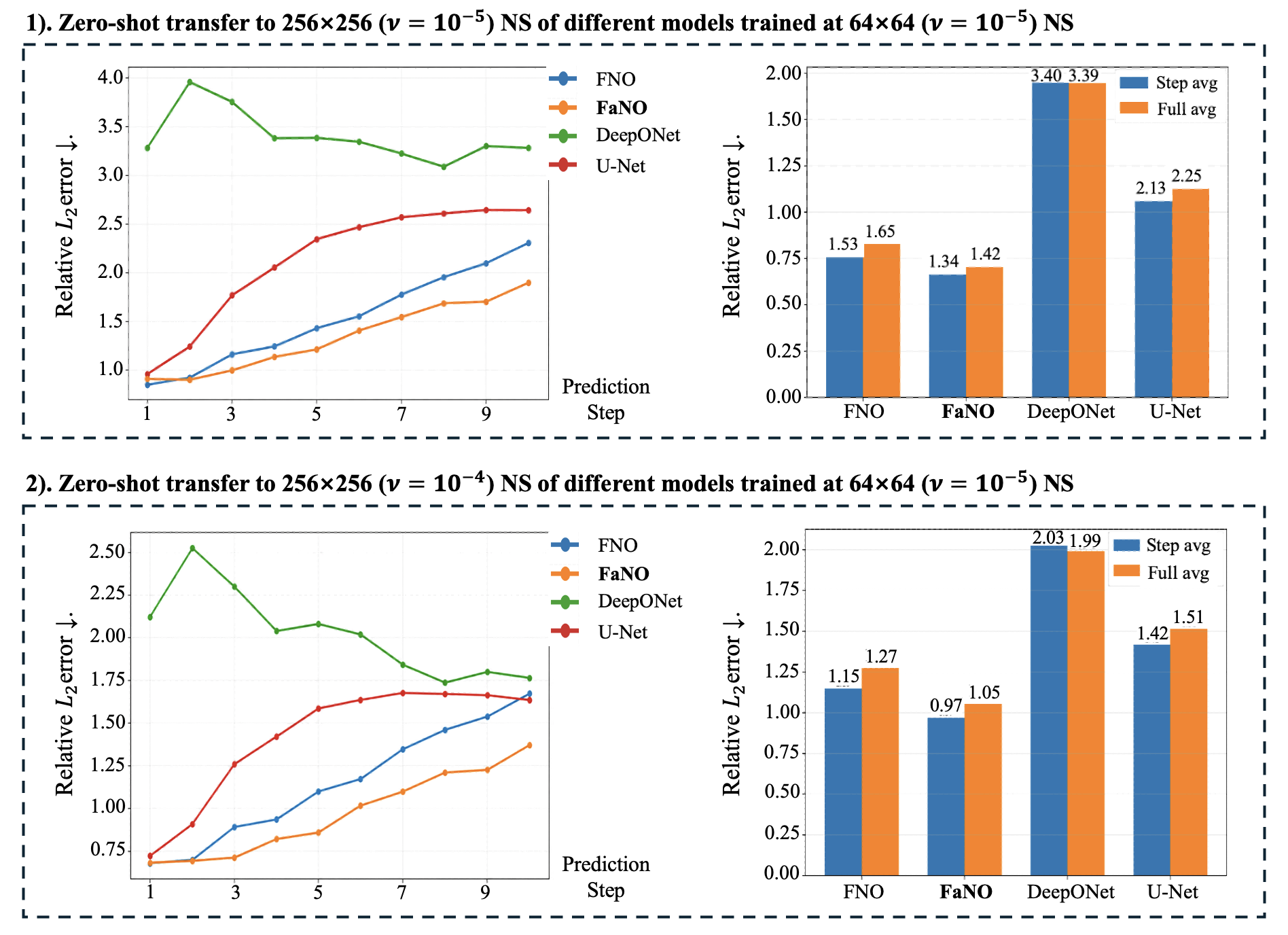}
\caption{Cross-scale and cross-regime rollout evaluation on $256 \times 256$ Navier--Stokes turbulence (NS).}
\label{fig:NS_256}
\end{figure*}

We further evaluate whether the advantages of response factorization transfer from controlled simulations to real-world Earth-system forecasting. We consider WeatherBench~\cite{rasp2020weatherbench}, an autoregressive global weather prediction benchmark constructed from the ERA5 reanalysis archive~\cite{hersbach2020era5}. See details in the Supplementary Information.

WeatherBench provides multivariate atmospheric fields at multiple spatial resolutions, including $5.625^\circ$ ($32\times64$), $2.8125^\circ$ ($64\times128$), and $1.40625^\circ$ ($128\times256$). Compared with SWE, this benchmark presents substantially greater modeling complexity due to the coexistence of local transient weather events, planetary-scale circulation structures, and strong multiscale coupling across atmospheric variables. We compare \ourmodel\ with representative climate modeling methods, including Transformer-based ClimaX~\cite{nguyen2023climax}, Fourier neural operator-based FourCastNet~\cite{pathak2022fourcastnet} and spherical neural operator-based SFNO~\cite{bonev2023spherical}.
Following prior works~\cite{nguyen2023climax,tang2026generalized}, we evaluate multiscale forecasting of all the models at 24, 72, and 120 hours (steps) across multiple spatial resolutions, including cross-resolution extrapolation at $128\times256$. Performance is measured using anomaly correlation coefficient (ACC). Details are in the Supplementary Information.

Across all evaluated atmospheric variables and spatial resolutions shown in Table.~\ref{tab:WB_train32x64} \&~\ref{tab:WB_train64x128}, \ourmodel~achieves the highest ACC at medium- and long-range forecasting horizons ($3$-day and $5$-day prediction) across resolutions, while remaining highly competitive at short-range forecasting and using only $2.7$M parameters, substantially fewer than competing methods. More importantly, \ourmodel~consistently outperforms single-branch operator (SFNO) under cross-resolution transfer, maintaining accurate prediction quality across unseen spatial discretizations in Table.~\ref{tab:WB_train64x128}.
The qualitative comparisons in Fig.~\ref{fig:weather} further highlight the advantages of response factorization in long-horizon forecasting. Existing methods accumulate visible structural distortions and phase errors, whereas \ourmodel~preserves coherent planetary-scale flow structures and local weather patterns with minimal degradation even at $120$h prediction.

These results suggest that the benefits of response factorization extend beyond controlled simulation benchmarks. By moving beyond purely equivariant formulations and explicitly separating dynamic and persistent responses ($\mathbf{S_E}+\mathbf{S_I}$), \ourmodel~provides a scalable operator design that remains consistent across temporal horizons, spatial resolutions, and forecasting regimes, bringing neural operator learning closer to real-world Earth-system deployment.

\subsection{Factorized operator stabilizes fluid dynamics across scales, regimes and geometries}

\begin{figure*}[hb]
\centering
\includegraphics[width=0.99\linewidth]{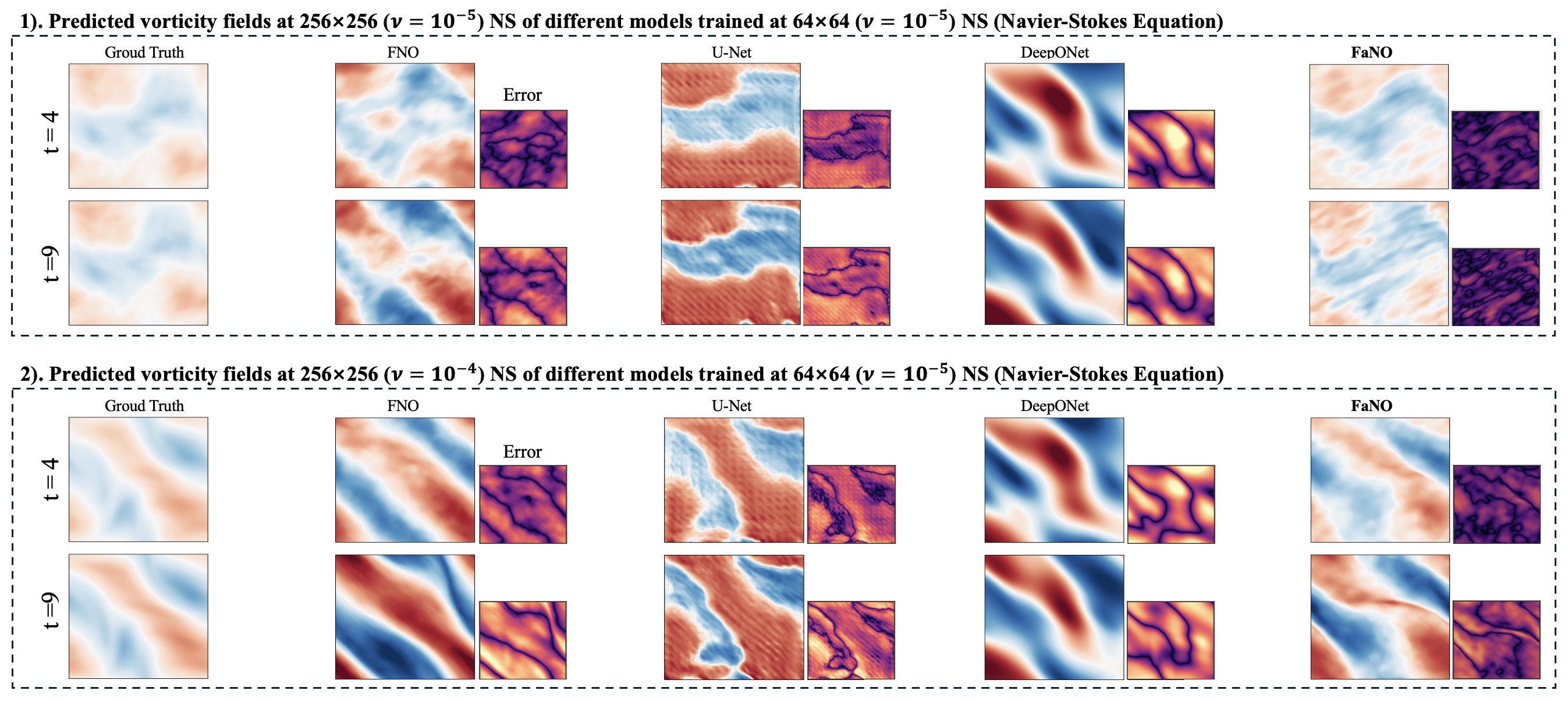}
\caption{Predicted Vorticity fields and the absolute error maps of different methods on Navier--Stokes turbulence.}
\label{fig:qualitative_NS}
\end{figure*}

We further investigate whether the advantages of response factorization extend beyond spherical geophysical systems. We first evaluate \ourmodel\ on a canonical fluid-dynamics setting: the two-dimensional incompressible Navier--Stokes equation~\cite{li2020fourier}. This benchmark serves as a standard testbed for autoregressive operator learning on Euclidean grids. We compare \ourmodel\ with representative methods on the regular grid, including U-Net~\cite{ronneberger2015u}, DeepONet~\cite{lu2019deeponet} and FNO~\cite{li2020fourier}. Following the standard FNO protocol~\cite{li2020fourier}, all models are trained at \(64\times64\) resolution with viscosity \(\nu=10^{-5}\), using the first 10 vorticity frames as input and autoregressively predicting the next 10 frames; we
 use $L_2$ error as the evaluation metric. Although the geometry is regular, the dynamics are strongly nonlinear and sensitive to recursive rollout errors. We therefore use this setting as a basic test of whether response factorization remains reliable when moving from spherical geophysical dynamics to fluid prediction on Euclidean grids, including viscosity transfer from \(\nu=10^{-5}\) to \(\nu=10^{-4}\), cross-resolution transfer from \(64\times64\) to \(256\times256\), and a combined setting involving simultaneous changes in resolution and viscosity. Details are provided in the Supplementary Information.

As shown in Fig.~\ref{fig:NS_64} and~\ref{fig:NS_256}, \ourmodel\ consistently achieves lower rollout errors than other methods, especially the single-branch FNO baseline. Meanwhile, \ourmodel\ uses only \(45.8\%\) of FNO's parameters (0.190M vs. 0.415M). The advantage is not restricted to the in-distribution setting: across viscosity transfer, cross-resolution transfer, and the combined resolution--viscosity shift, \ourmodel\ maintains lower errors than FNO, suggesting that the factorized representation is not tied to a single grid resolution or physical regime. The vorticity fields in Fig.~\ref{fig:qualitative_NS} further show how this improvement appears along the rollout trajectory. FNO captures the coarse vortex morphology but gradually accumulates phase shifts and structural distortions; U-Net introduces local artifacts, while DeepONet tends to produce overly smooth fields. In contrast, \ourmodel\ better preserves coherent vortex organization and keeps prediction errors more spatially concentrated, indicating that separating rapidly varying and persistent spectral responses provides a more stable inductive bias for autoregressive fluid prediction.

To test whether the same advantage persists in a more geometry-constrained fluid system, we evaluate \ourmodel\ on CylinderFlow, a vortex-shedding dataset over unstructured meshes~\cite{pfaff2021meshgraphnets}. Compared with regular-grid Navier--Stokes prediction, CylinderFlow introduces an explicit cylindrical obstacle: fixed boundary constraints interact with dynamically evolving downstream wakes, and the computational domain is non-rectangular. This setting therefore provides a stricter test of long-horizon autoregressive stability under both fluid dynamics and geometric constraints. Following the rollout protocol on CylinderFlow~\cite{pfaff2021meshgraphnets}, all models are evaluated on the same test trajectories with identical mesh inputs, spectral operator caches, normalization statistics, and rollout settings. We use mean squared error (MSE) as the evaluation metric for CylinderFlow rollout prediction. We compare \ourmodel\ with the single-branch dynamic baseline (NORM)~\cite{chen2024learning} and DiffusionNet~\cite{sharp2022diffusionnet} under long-horizon autoregressive prediction. Details are provided in the Supplementary Information.

\begin{figure*}[hb]
\centering
\includegraphics[width=0.95\linewidth]{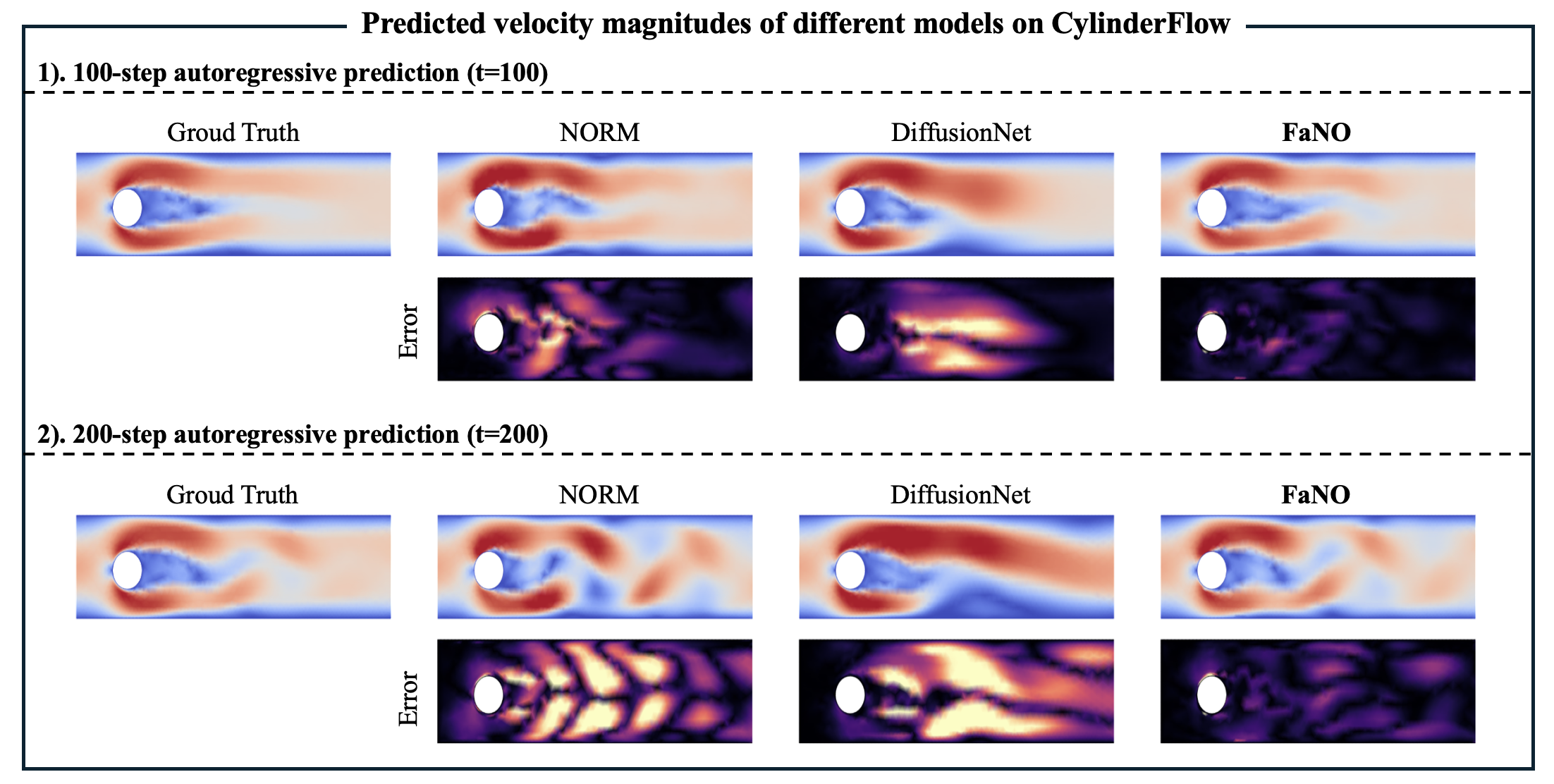}
\caption{Cylinder wake prediction and the absolute error maps on unstructured meshes.}
\label{fig:qualitative_Cylinder}
\end{figure*}

Fig.~\ref{fig:quanti_cylinder} reports two complementary CylinderFlow metrics: per-step rollout mean squared error (MSE) and trajectory-wise late-horizon averaged MSE, with specific definitions provided in the Supplementary Information. The per-step curves show that the performance gap becomes increasingly pronounced as the rollout horizon grows: NORM and DiffusionNet accumulate errors more rapidly under recursive prediction, whereas \ourmodel\ maintains lower error growth. At the final rollout step(\(t=200\)), \ourmodel\ reduces the per-step MSE by approximately \(28.8\%\) compared with the single-branch NORM baseline. The trajectory-wise late-horizon averaged MSE further shows that \ourmodel\ shifts the error distribution toward lower values, indicating improved stability across long-horizon test trajectories. The wake visualizations in Fig.~\ref{fig:qualitative_Cylinder} show the corresponding field-level behavior. NORM and DiffusionNet gradually accumulate visible distortions in the downstream wake region, leading to deviations from the ground-truth vortex-shedding structures during long rollout. In contrast, \ourmodel\ better preserves coherent downstream wake patterns and produces lower error maps at later steps.

Taken together, these results show a progressive generalization path: response factorization first improves autoregressive fluid prediction on the canonical regular-grid Navier--Stokes benchmark, and then remains effective in the more geometry-constrained CylinderFlow setting on unstructured meshes. This suggests that the factorized representation ($\mathbf{S_E}+\mathbf{S_I}$) is not limited to spherical geophysical systems or regular Euclidean grids, but provides a stable inductive bias for cross-scale physical generalization under temporal, spatial, parametric, and geometric shifts.

\begin{figure*}[hb]
\centering
\includegraphics[width=0.95\linewidth]{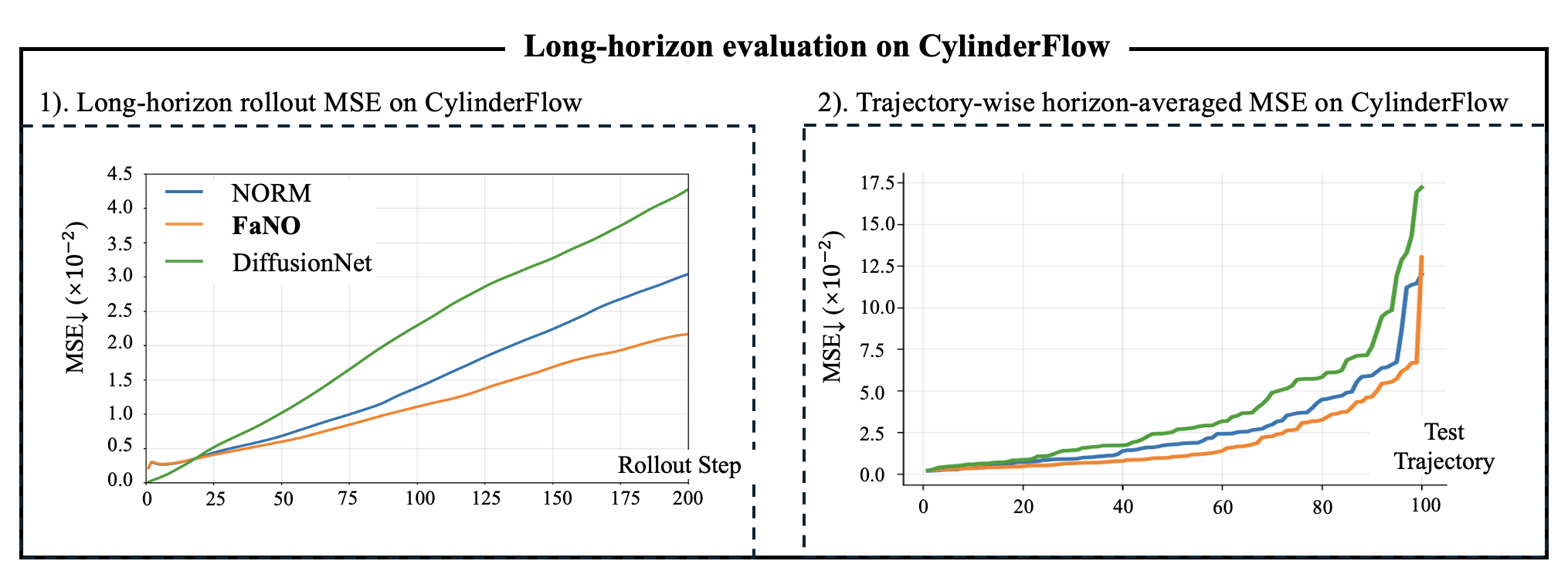}
\caption{Mean squared error (MSE) comparison of long-horizon rollout on the CylinderFlow.}
\label{fig:quanti_cylinder}
\end{figure*}

\subsection{Factorization extends beyond time-dependent physics}

\begin{figure*}[t]
\centering
\includegraphics[width=0.99\linewidth]{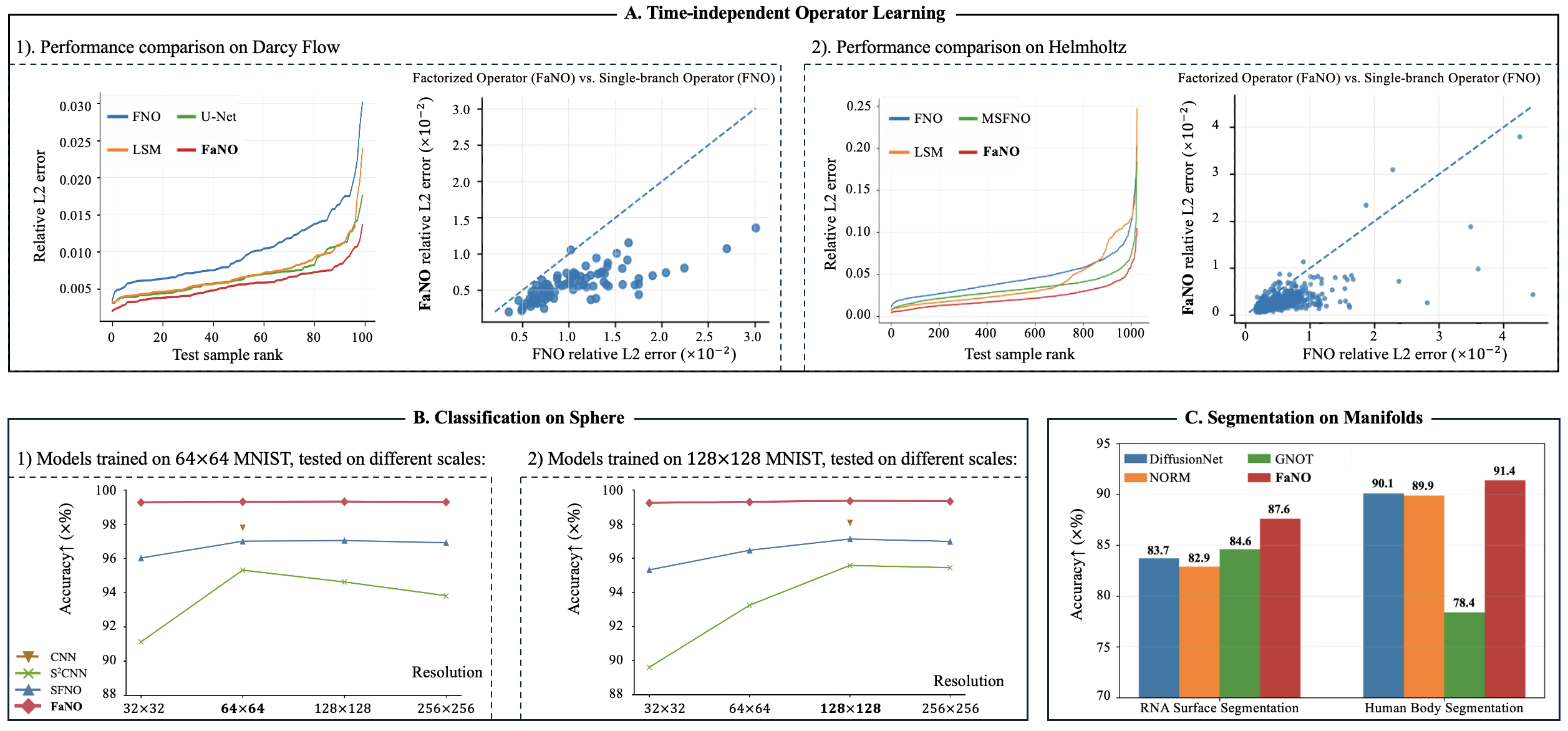}
\caption{Generalization beyond time-dependent physics.
\textbf{A}. Time-independent operator learning on Darcy flow and Helmholtz benchmarks. For each task, the sorted sample-wise relative \(L_2\) errors and the sample-wise comparison between \ourmodel\ and FNO are reported.
\textbf{B}. Digit classification accuracy on the sphere across spatial resolutions.
\textbf{C}. Segmentation accuracy on RNA molecular surfaces and human body meshes.}
\label{fig:generalization}
\end{figure*}

Finally, we examine whether response factorization remains useful beyond temporal physical modeling. We consider three complementary settings: time-independent operator learning, geometric classification and segmentation. These tasks differ in governing structure, domain geometry, and output type, but they share a common challenge: the model must separate stable global structure from input-dependent local variations. We therefore use them to test whether the proposed factorized representation provides a general modeling principle rather than an improvement restricted to fluid dynamics or temporal modeling.


We first evaluate elliptic and wave-type operator learning on Darcy flow and Helmholtz equations. The Darcy experiments follow the standard FNO benchmark protocol~\cite{li2020fourier}, while the Helmholtz experiments follow the benchmark setup of Subramanian~\etal~\cite{subramanian2023towards}. 
In both tasks, all models are trained under the same data split and optimization protocol, and performance is measured by the test relative \(L_2\) error. 
These two benchmarks stress complementary forms of Euclidean heterogeneity. Darcy flow requires inferring a globally coupled elliptic solution from discontinuous coefficients, whereas Helmholtz equations involve oscillatory responses that are sensitive to local phase and frequency variations. As shown in Fig.~\ref{fig:generalization}A, \ourmodel\ achieves lower errors than other representative methods on both tasks, including FNO~\cite{li2020fourier}, U-Net~\cite{ronneberger2015u}, LSM~\cite{wu2023solving} and MsFNO~\cite{you2025mscalefno}. More importantly, its sorted sample-wise relative \(L_2\) curves remain below those of competing methods across most test samples, with a clearer advantage in the high-error tail. This indicates that factorization is particularly helpful for difficult samples where heterogeneous coefficients or local oscillatory structures make a single spectral response less stable. The sample-wise scatter plots further show that \ourmodel\ improves over FNO on the majority of test cases, suggesting that the gain is distributed across the test set rather than concentrated on a few favorable examples.

We next study non-operator classification and segmentation on geometries. For Spherical MNIST, all models are trained and evaluated following the S$^2$CNN protocol~\cite{cohen2018spherical}. For RNA molecular surface and human body segmentation, all models follows the DiffusionNet protocol~\cite{sharp2022diffusionnet}. These tasks involve irregular geometry, non-uniform sampling, and substantial shape variation.
As shown in Fig.~\ref{fig:generalization}B, \ourmodel\ maintains consistently higher digital classification accuracy across different test resolutions, while competing methods show more visible degradation under resolution shifts. This suggests that response factorization also improves resolution-consistent representation learning on spherical domains.
Fig.~\ref{fig:generalization}C shows that \ourmodel\ achieves the highest surface segmentation accuracy among geometry-aware baselines, including DiffusionNet~\cite{sharp2022diffusionnet}, NORM~\cite{chen2024learning}, and GNOT~\cite{hao2023gnot}. This demonstrates that the benefit of factorization extends from structured fields to irregular geometric data.

Together, these experiments show that the advantages of response factorization ($\mathbf{S_E}+\mathbf{S_I}$) are not limited to physical or temporal-related modeling. Across Euclidean, spherical, and manifold domains, \ourmodel~improves accuracy, robustness on difficult samples, and parameter efficiency. These results support the view that decomposing operator responses into complementary components provides a scalable inductive bias for heterogeneous operator and representation learning.

%% file: Tables/WB_5deg_ACC.tex
\begin{table*}[hb]
    \caption{ACC↑ (in \%) on \textbf{32×64 WeatherBench} for six variables and their average at 1, 3, and 5 days. Rows grouped by forecast horizon. \textbf{Bold} indicates best performance in each row group.
    }
    \label{tab:WB_train32x64}
    \setlength{\tabcolsep}{7pt}
    \def\arraystretch{1.2}
    \resizebox{1.0\textwidth}{!}{

    
    
    \begin{tabular}{l!{\vrule}c!{\vrule}c!{\vrule}cccccc!{\vrule}c}    \toprule
    
\multirow{2}{*}{\textbf{Method}} 
& \multirow{2}{*}{
\makecell[c]{\textbf{Grid-}\\\textbf{invariant?}}}   
& \multirow{2}{*}{\textbf{Params}}
& \multicolumn{6}{c!{\vrule}}{\textbf{Prediction Variables}} 
& \multirow{2}{*}{\textbf{Average}} \\
\cmidrule(lr){4-9}
& & & 2T & 10U & 10V & U600 & V600 & T600 \\
    \midrule
     \multicolumn{9}{c}{\textbf{1 Day Forecast}} \\
     \cmidrule(lr){1-10}
     ClimaX~\cite{nguyen2023climax}    & \ding{55} & 5.4 M
     & 96.1 ± 2.07 
     & \textbf{92.0 ± 0.72} 
     & \textbf{90.9 ± 0.80} 
     & \textbf{92.3 ± 0.69} 
     & \textbf{90.8 ± 0.85} 
     & 96.0 ± 1.63
     & \textbf{93.0} \\
    FourCastNet~\cite{pathak2022fourcastnet}     & \ding{55}  & 5.3 M 
    & 96.1 ± 2.15 
    & 90.2 ± 0.83 
    & 89.8 ± 0.91 
    & 92.1 ± 0.79 
    & 90.5 ± 0.92 
    & 95.7 ± 1.57 
    & 92.4 \\
    SFNO~\cite{bonev2023spherical}       & \ding{51} & 3.7 M
    & 95.3 ± 2.23 
     & 85.5 ± 1.65 
     & 84.2 ± 1.60 
     & 87.1 ± 1.38 
     & 84.9 ± 1.79 
     & 94.7 ± 2.08 
     & 88.6  \\
    \ournetwork~(Ours)    & \ding{51} & 2.7 M
    & \textbf{96.6 ± 2.02}
    & 88.5 ± 1.07 
    & 88.6 ± 1.16 
    & 90.7 ± 0.82 
    & 89.8 ± 1.08 
    & \textbf{96.3 ± 1.60}
    & 91.8 \\
    
    \cmidrule(lr){1-10}
    \multicolumn{9}{c}{\textbf{3 Day Forecast}} \\
    \cmidrule(lr){1-10}
     ClimaX~\cite{nguyen2023climax}       & \ding{55}  & 5.4 M    
     & 90.2 ± 5.63 
     & 57.8 ± 5.12 
     & 56.4 ± 5.08 
     & 66.5 ± 4.65 
     & 57.7 ± 5.26 
     & 83.5 ± 8.01 
     & 68.7 \\
    FourCastNet~\cite{pathak2022fourcastnet}    & \ding{55}  & 5.3 M  
    & 88.9 ± 6.47 
    & 56.6 ± 4.72 
    & 54.5 ± 4.66 
    & 65.9 ± 4.25 
    & 55.4 ± 5.52 
    & 82.3 ± 8.38 
    & 67.3 \\
    SFNO~\cite{bonev2023spherical}       & \ding{51} & 3.7 M
    & 91.0 ± 5.18 
    & 59.1 ± 5.02 
    & 55.9 ± 5.11 
    & 68.3 ± 4.49 
    & 58.1 ± 5.35 
    & 83.7 ± 8.08 
    & 69.4 \\
    \ournetwork~(Ours)    & \ding{51} & 2.7 M
    & \textbf{92.2 ± 4.70}
    & \textbf{63.3 ± 4.52} 
    & \textbf{61.6 ± 4.61}
    & \textbf{71.2 ± 4.22} 
    & \textbf{63.7 ± 4.97} 
    & \textbf{85.4 ± 7.33}
    & \textbf{72.9} \\
      
    \cmidrule(lr){1-10}
    \multicolumn{9}{c}{\textbf{5 Day Forecast}} \\
    \cmidrule(lr){1-10}
    ClimaX~\cite{nguyen2023climax}        & \ding{55}   & 5.4 M 
    & 84.6 ± 9.27 
    & 28.5 ± 9.44 
    & 21.5 ± 8.23 
    & 35.6 ± 9.89 
    & 14.4 ± 8.64 
    & 66.7 ± 18.3 
    & 41.9 \\
     FourCastNet~\cite{pathak2022fourcastnet} & \ding{55} & 5.3 M  
     & 83.5 ± 10.8 
     & 29.4 ± 8.74 
     & 22.8 ± 7.67 
     & 36.4 ± 10.1 
     & 14.1 ± 7.92 
     & 64.7 ± 19.6 
     & 41.8 \\
    SFNO~\cite{bonev2023spherical}       & \ding{51} & 3.7 M
    & 87.2 ± 7.65
     & 37.6 ± 8.70 
     & 30.5 ± 9.15 
     & 45.6 ± 9.21 
     & 26.6 ± 9.48 
     & 70.9 ± 16.3 
     & 49.7 \\
    \ournetwork~(Ours)    & \ding{51} & 2.7 M
    & \textbf{88.0 ± 7.32}
    & \textbf{42.0 ± 8.10}
    & \textbf{35.3 ± 8.18}
    & \textbf{50.5 ± 8.22}
    & \textbf{32.3 ± 8.54}
    & \textbf{73.6 ± 14.2}
    & \textbf{53.6} \\
   
    \bottomrule
    \end{tabular}
    }
    \vspace{0.3cm}
\end{table*}

%% file: Tables/WB_2deg_ACC.tex
\newcommand{\tightbox}[1]{\begingroup\setlength{\fboxsep}{1pt}\colorbox{gray!20}{#1}\endgroup}

\begin{table*}[h]
    \caption{ACC↑ (in \%) of different models trained on \textbf{64×128 WeatherBench} for six variables and their average at 1, 3, and 5 days. Rows grouped by scales (forecast horizon and resolution). \textbf{Bold} indicates best performance in each row group.
    }
    \label{tab:WB_train64x128}
    \setlength{\tabcolsep}{7pt}
    \def\arraystretch{1.2}
    \resizebox{1.0\textwidth}{!}{

    
    
    \begin{tabular}{l!{\vrule}c!{\vrule}c!{\vrule}cccccc!{\vrule}c}    \toprule
    
\multirow{2}{*}{\textbf{Method}} & \multirow{2}{*}{\makecell[c]{\textbf{Grid-}\\\textbf{invariant?}}}   & \multirow{2}{*}{\textbf{Params}}
& \multicolumn{6}{c!{\vrule}}{\textbf{Prediction Variables}} 
& \multirow{2}{*}{\textbf{Average}} \\
\cmidrule(lr){4-9}
& & & 2T & 10U & 10V & U600 & V600 & T600 \\
    \midrule
     \multicolumn{9}{c}{\textbf{1 Day Forecast (Tested on $64 \times 128$ WB)}} \\
     \cmidrule(lr){1-10}
     ClimaX~\cite{nguyen2023climax}    & \ding{55} & 7.0 M
     & 96.4 ± 2.24
    & \textbf{93.0 ± 0.72}
    & \textbf{92.6 ± 0.87}
    & \textbf{93.9 ± 0.75}
    & \textbf{92.5 ± 0.88}
    & 96.8 ± 1.56
    & \textbf{94.2} \\
    FourCastNet~\cite{pathak2022fourcastnet}     & \ding{55}  & 5.9 M 
    & 95.9 ± 2.32 
    & 91.8 ± 0.73 
    & 91.6 ± 0.77
    & 93.1 ± 0.70 
    & 91.9 ± 0.92 
    & 96.2 ± 1.63 
    & 93.4 \\
    SFNO~\cite{bonev2023spherical}       & \ding{51} & 3.7 M
    & 95.3 ± 2.45 
    & 89.4 ± 1.28  
    & 88.5 ± 1.33
    & 90.8 ± 1.25
    & 89.2 ± 1.20
    & 95.0 ± 2.17
    & 91.4 \\
    \ournetwork~(Ours)    & \ding{51} & 2.7 M
    & \textbf{97.2 ± 1.68}
    & 91.9 ± 0.74 
    & 91.7 ± 0.79 
    & 92.8 ± 0.75 
    & 91.9 ± 0.82 
    & \textbf{97.0 ± 1.27}
    & 93.7 \\
    
    \cmidrule(lr){1-10}
    \multicolumn{9}{c}{\textbf{3 Day Forecast (Tested on $64 \times 128$ WB)}} \\
    \cmidrule(lr){1-10}
     ClimaX~\cite{nguyen2023climax}       & \ding{55} & 7.0 M    
    & 89.9 ± 6.02
    & 64.6 ± 4.04
    & 63.2 ± 4.08
    & 70.0 ± 3.65
    & 65.8 ± 4.49
    & 86.1 ± 6.44
    & 73.3 \\
    FourCastNet~\cite{pathak2022fourcastnet}    & \ding{55}  & 5.9 M  
    & 88.8 ± 6.32
    & 62.2 ± 4.65
    & 60.0 ± 4.72
    & 69.5 ± 4.18
    & 62.3 ± 5.51
    & 85.5 ± 7.06
    & 71.5 \\
    SFNO~\cite{bonev2023spherical}       & \ding{51} & 3.7 M
    & 90.3 ± 4.95
    & 64.2 ± 4.61
    & 62.7 ± 4.63
    & 70.6 ± 3.92
    & 64.9 ± 4.87
    & 86.3 ± 6.75
    & 73.2 \\
    \ournetwork~(Ours)    & \ding{51} & 2.7 M
    & \textbf{92.9 ± 3.90}
    & \textbf{70.5 ± 3.21} 
    & \textbf{69.4 ± 3.45}
    & \textbf{77.3 ± 2.77}
    & \textbf{72.6 ± 3.58}
    & \textbf{88.8 ± 4.93} 
    & \textbf{78.6} \\

%

    \cmidrule(lr){1-10}
    \multicolumn{9}{c}{\textbf{5 Day Forecast (Tested on $64 \times 128$ WB)}} \\
    \cmidrule(lr){1-10}
    ClimaX~\cite{nguyen2023climax}        & \ding{55} & 7.0 M 
    & 83.7 ± 9.92
    & 30.6 ± 8.58
    & 24.8 ± 8.28
    & 38.7 ± 9.63
    & 19.3 ± 8.75
    & 68.1 ± 17.5
    & 44.2 \\
     FourCastNet~\cite{pathak2022fourcastnet} & \ding{55} & 5.9 M  
    & 83.1 ± 10.6
    & 30.5 ± 8.72
    & 23.5 ± 6.68
    & 37.7 ± 10.3
    & 14.2 ± 7.19
    & 66.0 ± 19.0
    & 42.5 \\
     SFNO~\cite{bonev2023spherical}       & \ding{51} & 3.7 M
    & 86.3 ± 8.92
    & 39.4 ± 8.38
    & 33.2 ± 9.11
    & 46.8 ± 8.82
    & 27.2 ± 9.96
    & 72.3 ± 14.8
    & 50.9 \\
    \ournetwork~(Ours)    & \ding{51} & 2.7 M
    & \textbf{88.2 ± 6.89}
    & \textbf{44.7 ± 7.55}
    & \textbf{38.5 ± 8.27}
    & \textbf{53.1 ± 7.20}
    & \textbf{37.6 ± 7.56}
    & \textbf{74.5 ± 12.8} 
    & \textbf{56.1} \\

    \midrule
     \multicolumn{9}{c}{\textbf{1 Day Forecast (Cross-resolution generalized on $128 \times 256$ WB)}} \\
     \cmidrule(lr){1-10}
    SFNO~\cite{bonev2023spherical}       & \ding{51} & 3.7 M
    & 95.0 ± 2.52
    & 88.5 ± 1.30
    & 87.8 ± 1.51
    & 90.3 ± 1.28
    & 88.0 ± 1.44
    & 94.8 ± 2.19
    & 90.7 \\
    \ournetwork~(Ours)    & \ding{51} & 2.7 M
    & \textbf{96.8 ± 1.77}
    & \textbf{90.6 ± 0.78}
    & \textbf{90.8 ± 0.81} 
    & \textbf{92.5 ± 0.80}
    & \textbf{91.8 ± 0.85}
    & \textbf{96.7 ± 1.33}
    & \textbf{93.2} \\
    
    \cmidrule(lr){1-10}
    \multicolumn{9}{c}{\textbf{3 Day Forecast (Cross-resolution generalized on $128 \times 256$ WB)}} \\
    \cmidrule(lr){1-10}
    SFNO~\cite{bonev2023spherical}       & \ding{51} & 3.7 M
    & 90.1 ± 4.98
    & 63.6 ± 4.67
    & 62.2 ± 4.72
    & 70.3 ± 4.05
    & 63.8 ± 5.21
    & 86.0 ± 6.82
    & 72.7 \\
    \ournetwork~(Ours)    & \ding{51} & 2.7 M
    & \textbf{92.8 ± 3.95}
    & \textbf{70.1 ± 3.32}
    & \textbf{69.2 ± 3.50}
    & \textbf{77.1 ± 2.79}
    & \textbf{72.4 ± 3.77}
    & \textbf{88.7 ± 4.95}
    & \textbf{78.4} \\

%

    \cmidrule(lr){1-10}
    \multicolumn{9}{c}{\textbf{5 Day Forecast (Cross-resolution generalized on $128 \times 256$ WB)}} \\
    \cmidrule(lr){1-10}
     SFNO~\cite{bonev2023spherical}       & \ding{51} & 3.7 M
    & 86.0 ± 9.10
    & 38.5 ± 8.43
    & 32.4 ± 9.24
    & 46.2 ± 8.89
    & 25.8 ± 12.1
    & 72.1 ± 15.5
    & 50.2 \\
    \ournetwork~(Ours)    & \ding{51} & 2.7 M
    & \textbf{88.1 ± 7.09}
    & \textbf{44.1 ± 7.58}
    & \textbf{37.9 ± 8.38}
    & \textbf{52.8 ± 7.42}
    & \textbf{37.2 ± 8.72}
    & \textbf{74.4 ± 13.6}
    & \textbf{55.8} \\
   
    \bottomrule
    \end{tabular}
    }
\end{table*}